\documentclass[runningheads]{llncs}

\usepackage{eccv}

\usepackage{eccvabbrv}

\usepackage{graphicx}
\usepackage{booktabs}

\usepackage[accsupp]{axessibility}  

\definecolor{eccvblue}{rgb}{0.21,0.49,0.74}
\usepackage[pagebackref,breaklinks,colorlinks,allcolors=eccvblue]{hyperref}

\usepackage{orcidlink}

\usepackage{bm}
\usepackage{booktabs}
\usepackage{multirow}
\usepackage{microtype}
\usepackage{epsfig}
\usepackage{caption}
\usepackage{float}
\usepackage{placeins}
\usepackage{color, colortbl}
\usepackage{stfloats}
\usepackage{enumitem}
\usepackage{tabularx}
\usepackage{xstring}
\usepackage{xspace}
\usepackage{url}
\usepackage{subcaption}       
\usepackage[dvipsnames]{xcolor}
\usepackage{ulem}

\newcommand\blfootnote[1]{%
  \begingroup
  \renewcommand\thefootnote{}\footnote{#1}%
  \addtocounter{footnote}{-1}%
  \endgroup
}

\definecolor{gtable}{rgb}{0.0, 0.5, 0.0}

\newcommand{\ra}[1]{\renewcommand{\arraystretch}{#1}}

\makeatletter

\newcommand{\Rmnum}[1]{\expandafter\@slowromancap\romannumeral #1@}
\makeatother

\usepackage[utf8]{inputenc}
\usepackage{amssymb}
\usepackage{bbding}
\usepackage{pifont}
\usepackage{wasysym}
\usepackage{utfsym}
\usepackage{fontawesome}
\definecolor{gray}{gray}{0.8}  

\definecolor{top}{HTML}{E7F0DC}          
\definecolor{baseline}{HTML}{EEEEEE}        
\definecolor{close_source}{HTML}{CBF1F5}   
\definecolor{open_source}{HTML}{FDE7BB}       
\definecolor{baseline}{HTML}{EEEEEE}

\newcommand{\green}[1]{{\color[HTML]{5F8D4E}#1}}

\definecolor{gain}{HTML}{34a853}   

\definecolor{lost}{HTML}{ea4335}

\definecolor{my_red}{HTML}{FF0000}       
\definecolor{my_purple}{HTML}{AA96DA} 
\definecolor{my_orange}{HTML}{F07B3F} 

\definecolor{my_box_red}{HTML}{FFB4B4}   
\definecolor{my_box_purple}{HTML}{D3CEDF} 
\definecolor{my_box_orange}{HTML}{FFC3A1}

\begin{document}

\title{WeEdit: A Dataset, Benchmark and Glyph-Guided Framework for Text-centric Image Editing}

\author{
    Hui Zhang$^{1,2,3}$\quad
    Juntao Liu$^{3}$\quad
    Zongkai Liu$^{3}$\quad
    Liqiang Niu$^{3}$\quad 
    Fandong Meng$^{3,\dagger}$ \qquad  \\
    Zuxuan Wu$^{1,2,\dagger}$\quad
    Yu-Gang Jiang$^{1,2,\dagger}$
}

\institute {
$^{1}$Institute of Trustworthy Embodied AI, Fudan University \\
$^{2}$Shanghai Key Laboratory of Multimodal Embodied AI \quad
$^{3}$Weixin AI, Tencent 
\url{https://huizhang0812.github.io/WeEdit/}}

\titlerunning{WeEdit}


\authorrunning{H. Zhang et al.}


\maketitle

\vspace{-2.75em}
\begin{figure}[htbp] %
     \centering
     \includegraphics[width=1.0\textwidth]{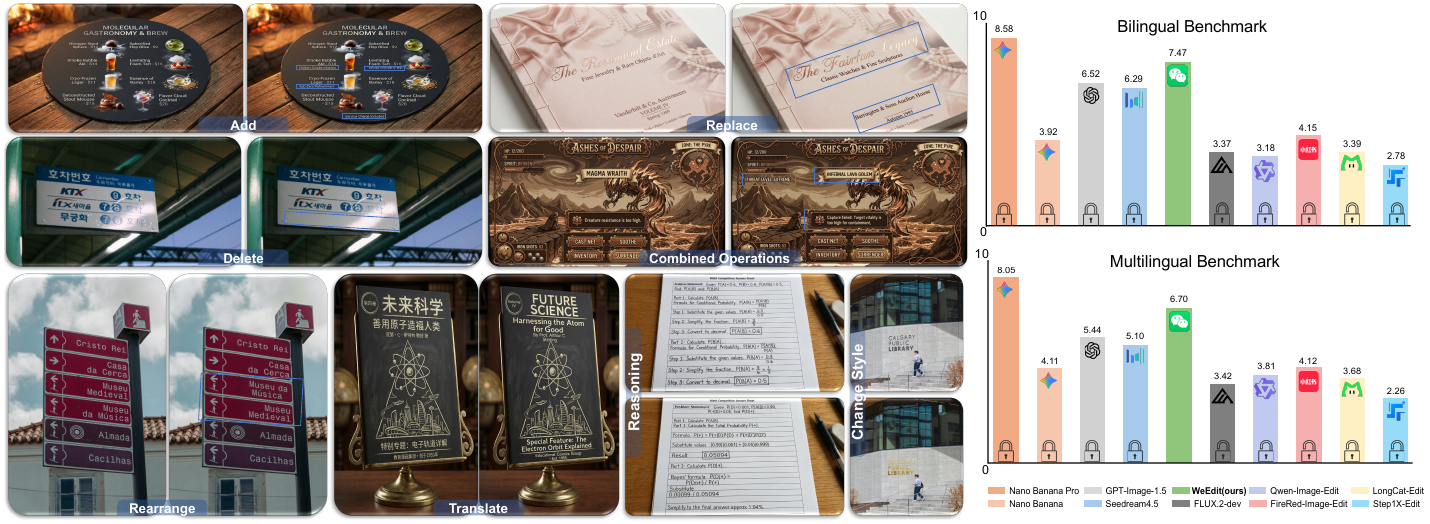}
    \vspace{-2em}
    \caption{\textbf{Left:} WeEdit achieves precise manipulation of textual content within images across diverse editing operations (edited regions are highlighted with \textcolor{eccvblue}{blue} bounding boxes). \textbf{Right:} WeEdit achieves the best performance among all open-source models on both bilingual and multilingual benchmarks, surpassing most proprietary models and ranking second only to Nano Banana Pro.}
    \label{fig:teaser}
\end{figure}

\vspace{-3.0em}
\begin{abstract}

Instruction-based image editing aims to modify specific content within existing images according to user-provided instructions while preserving non-target regions. 
Beyond traditional object- and style-centric manipulation, text-centric image editing focuses on modifying, translating, or rearranging textual elements embedded within images. 
However, existing leading models often struggle to execute complex text editing precisely, frequently producing blurry or hallucinated characters. 
We attribute these failures primarily to the lack of specialized training paradigms tailored for text-centric editing, as well as the absence of large-scale datasets and standardized benchmarks necessary for a closed-loop training and evaluation system. 
To address these limitations, we present WeEdit, a systematic solution encompassing a scalable data construction pipeline, two benchmarks, and a tailored two-stage training strategy. 
Specifically, we propose a novel HTML-based automatic editing pipeline, which generates 330K training pairs covering diverse editing operations and 15 languages, accompanied by standardized bilingual and multilingual benchmarks for comprehensive evaluation. 
On the algorithmic side, we employ glyph-guided supervised fine-tuning to inject explicit spatial and content priors, followed by a multi-objective reinforcement learning stage to align generation with instruction adherence, text clarity, and background preservation. 
Extensive experiments demonstrate that WeEdit outperforms previous open-source models by a clear margin across diverse editing operations.

\vspace{-0.75em}
\blfootnote{ $^{{\dagger}}$ Corresponding author.}
\end{abstract}

\section{Introduction}
\label{sec:intro}

In recent years, diffusion models~\cite{ho2020ddpm,dhariwal2021diffusionbeatgans,song2021ddim} have made significant progress in generative tasks, particularly in image generation and editing. Text-to-image models~\cite{rombach2022stablediffusion,esser2024sd3,flux,podell2023sdxl} aim to generate high-quality, visually appealing images from textual prompts. Upon this, image editing models~\cite{hertz2023prompt-to-prompt,brooks2023instructpix2pix,zhang2023magicbrush} 
have been developed to modify specific content within existing images based on editing instructions, while preserving non-target regions.

Currently, leading proprietary models (\eg Gemini-3-Pro-Image~\cite{google2025gemini-3.0-pro-image} and GPT-Image-1.5~\cite{openai2025gpt-image-1.5}) and open-source models (\eg FLUX.2-dev~\cite{flux2} and Qwen-Image-Edit~\cite{wu2025qwenimage}) excel in general image editing, demonstrating precise adherence to user instructions for object manipulation and style transfer. 
However, a critical yet fundamentally distinct dimension--\emph{text-centric image editing}, which involves modifying, translating, or rearranging textual elements embedded within images--remains largely underexplored. Such capabilities are important in a wide range of real-world applications, including infographic updating, poster modification, multilingual localization of user interfaces, and document editing.
Unlike previous editing tasks, this task requires models to possess accurate text recognition, precise layout planning, and clear text generation abilities, all while strictly preserving the background context (as shown in \cref{fig:teaser}).

These practical requirements expose the significant limitations of current paradigms. Existing models often struggle to follow complex editing instructions and tend to produce blurry or hallucinated characters, with performance deteriorating further when handling non-Latin scripts (\eg Arabic, Thai, and Hindi). We attribute these failures to critical gaps in current algorithms, training data, and evaluation methodologies:
\textbf{\Rmnum{1}) Algorithmic limitations:} Conventional image editing methods lacks specialized training paradigms explicitly designed to handle the unique text content modification within images.
\textbf{\Rmnum{2}) Data scarcity:} There is no large-scale, high-quality datasets specifically curated for diverse text-centric editing operations, particularly for multilingual settings.
\textbf{\Rmnum{3}) Evaluation gap:} The absence of a comprehensive, standardized benchmark tailored to this specific task hinders systematic model comparison, slowing down progress in this field.

To this end, we propose \textbf{WeEdit}, a systematic solution for text-centric image editing that jointly addresses the aforementioned challenges of model capability, training strategy, data scarcity, and evaluation standardization:
\textbf{\Rmnum{1}) Glyph-Guided Supervised Fine-Tuning:} To tackle the core difficulty of precise text placement and character-level accuracy, we introduce a text-aware fine-tuning approach. WeEdit first predicts the approximate position and scale of the target text, then renders a glyph layout as an explicit spatial prior to condition the diffusion process, granting direct spatial control and significantly improving glyph fidelity.
\textbf{\Rmnum{2}) Multi-Objective Reinforcement Learning:} To bridge the gap between pixel-level supervision and human-centric quality goals such as legibility and contextual coherence, we introduce a reinforcement learning stage with a reward function that jointly balances instruction adherence, text readability, and preservation of non-edited regions, leading to higher editing quality.
\textbf{\Rmnum{3}) Scalable Data Construction Pipeline:} We develop a scalable, HTML-based data construction pipeline that automatically synthesizes diverse editing pairs. The pipeline naturally extends to multilingual settings via a ``translate-then-edit'' workflow, yielding a large-scale training dataset covering a broad spectrum of editing operations and languages.
\textbf{\Rmnum{4}) Standardized Multilingual Benchmark:} We establish a comprehensive benchmark encompassing diverse editing operations with both Chinese--English bilingual data and multilingual evaluation across 15 widely used languages. We further design a multi-faceted evaluation scheme to ensure practical and reliable comparisons between the models.
Finally, as illustrated on the right side of \cref{fig:teaser}, our model outperforms existing open-source models and most proprietary counterparts, achieving SOTA performance that is second only to Gemini-3-Pro-Image (\ie Nano Banana Pro~\cite{google2025gemini-3.0-pro-image}).

\section{Related Work}
\label{sec:related work}
\vspace{-0.75em}
\subsection{General Image Generation and Editing}
\vspace{-0.5em}
Text-to-image generation~\cite{rombach2022stablediffusion,podell2023sdxl,saharia2022imagen,chen2023pixartalpha,li2024hunyuandit,peebles2023dit,gao2024lumina,liu2024playgroundv3,esser2024sd3,zheng2024cogview3,flux,cai2025hidream,gao2025seedream3.0,wu2026visual} aims to generate visual content from textual descriptions.
Image editing~\cite{brooks2023instructpix2pix,zhang2023magicbrush,yu2025anyedit,zhao2024ultraedit,labs2025flux,xiao2025omnigen,zhang2025icedit,lin2025uniworld,wu2025chronoedit,deng2025bagel,wu2025omnigen2} extends this capability by modifying targeted regions of an existing image according to user instructions while preserving non-targeted
content.
Recently, leading proprietary~\cite{google2025gemini-3.0-pro-image,openai2025gpt-image-1.5} and open-source models~\cite{cui2025emu3.5,li2025uniworldv2,flux2,wu2025qwenimage,team2025longcat,liu2025step1x,cao2025hunyuanimage,team2026firered}, have demonstrated strong performance on object modification and style transfer.
However, text-centric image editing---an equally critical yet
fundamentally distinct dimension that involves modifying, translating,
or rearranging textual elements embedded within images---remains
largely underexplored. When confronted with such tasks, existing models fail to follow editing instructions accurately and produce blurred or misspelled characters.
In this paper, we aim to systematically benchmark the text-centric editing capabilities of current models and propose effective methods
to bridge this gap.

\vspace{-0.75em}
\subsection{Text-aware Image Generation and Editing}
\vspace{-0.5em}
To improve text rendering quality in generated images, recent approaches incorporate additional information such as character-level linguistic features~\cite{liu2024Glyph-byt5,liu2024Glyph-byt5-v2,zhao2024udifftext}, explicit character bounding boxes~\cite{chen2024textdiffuser2,zhang2025creatilayout,zhang2025creatidesign,zhou2025dreamrenderer,zhou2024migc,zhou2024migc++}, and auxiliary visual conditions like rendered glyph images~\cite{chen2023textdiffuser,tuo2023anytext,yang2023glyphcontrol,ma2023glyphdraw,zhang2025creatiposter}.
While these approaches have improved text fidelity in image generation, few have focused on the editing task.
In this work, we propose to introduce glyph images as explicit spatial priors into text-centric image editing, and leveraging the understanding and planning capabilities of Vision-Language Models to automatically generate glyph images of target texts.

\vspace{-0.75em}
\subsection{Datasets and Benchmarks for Image Editing}
\vspace{-0.5em}
Existing image editing datasets~\cite{zhang2023magicbrush,zhao2024ultraedit,hui2024hqedit,ye2025echo4o,wei2024omniedit,wang2025gpt-iamge-edit,qian2025pico-banana,ma2025x2edit,chen2025sharegpt-4o,chen2025opengpt} and benchmarks~\cite{liu2025step1x,ye2025imgedit,wang2025i2i,han2025unireditbench,chang2025bytemorph,zhao2025risebench} are primarily designed for general-purpose editing scenarios.
Some works involve text-related editing tasks~\cite{team2026firered,ma2025x2edit}, but they are limited in terms of operation diversity, language coverage, and evaluation granularity. Thus, we propose a novel automated data construction pipeline as well as the first comprehensive benchmark tailored for text-centric image editing.

\vspace{-0.75em}
\subsection{Post-training for Diffusion Models}
\vspace{-0.5em}
Parameter-efficient supervised fine-tuning (SFT), such as LoRA~\cite{hu2022lora}, has become a mainstream approach for adapting pre-trained diffusion models to downstream tasks or for incorporating auxiliary control signals~\cite{tan2025ominicontrol,huang2024context,ruiz2023dreambooth,wu2025UNO}. 
Reinforcement learning (RL)-based~\cite{sutton2018reinforcement} post-training further aligns generation with human preferences~\cite{kirstain2023pick,xu2023imagereward,zheng2025diffusionnft,liu2025flowgrpo,li2025mixgrpo,xue2025dancegrpo,wu2025editreward}. 
In this paper, we present a dedicated post-training framework for text-centric image editing, which consists of glyph-guided SFT to inject explicit visual priors, followed by a tailored reward design in the RL stage to ensure editing accuracy.

\vspace{-0.75em}
\section{Method}
\label{sec:method}
\vspace{-0.5em}
In this section, we first review the preliminaries of diffusion and flow-matching models as well as the formulation of image editing (\cref{sec:preliminary}). We then present our two-stage training framework: a glyph-guided supervised fine-tuning stage that injects explicit spatial and content priors for accurate text rendering (\cref{sec:sft}), followed by a reinforcement learning stage that optimizes human-centric objectives such as instruction adherence, text clarity, and background preservation (\cref{sec:rlhf}).

\vspace{-0.75em}
\subsection{Preliminary}
\label{sec:preliminary}
\vspace{-0.25em}
\paragraph{\textbf{Diffusion and Flow-Matching Models.}}
Diffusion models generate data by reversing a forward noising process that gradually corrupts clean data $\mathbf{x}_0 \sim p_{\text{data}}$ with Gaussian noise. The noisy sample at timestep $t$ is obtained via reparameterization:
{
\setlength{\abovedisplayskip}{2pt}
\setlength{\belowdisplayskip}{3pt}
\begin{equation}
    \mathbf{x}_t = \alpha_t \mathbf{x}_0 + \sigma_t \boldsymbol{\epsilon}, \quad \boldsymbol{\epsilon} \sim \mathcal{N}(\mathbf{0}, \mathbf{I}),
\end{equation}
}
where $\alpha_t, \sigma_t$ define the noise schedule. Under the velocity parameterization, a neural network $\mathbf{v}_\theta(\mathbf{x}_t, t)$ is trained to predict the trajectory tangent by minimizing:
{
\setlength{\abovedisplayskip}{2pt}
\setlength{\belowdisplayskip}{2pt}
\begin{equation}
\label{eq:velocity_loss}
    \mathcal{L}_{\text{flow}} = \mathbb{E}_{t, \mathbf{x}_0, \epsilon}\left[ \| \mathbf{v}_\theta(\mathbf{x}_t, t) - \mathbf{v}_t \|_2^2 \right],
\end{equation}
}
where $\mathbf{v}_t = \frac{d\mathbf{x}_t}{dt} = \frac{d\alpha_t}{dt}\mathbf{x}_0 + \frac{d\sigma_t}{dt}\boldsymbol{\epsilon}$.
Samples are then generated by solving the ODE $\mathrm{d}\mathbf{x}_t / \mathrm{d}t = \mathbf{v}_\theta(\mathbf{x}_t, t)$ from $t{=}1$ (noise) to $t{=}0$ (data).
Rectified flow~\cite{liu2023flow} sets $\alpha_t = 1 - t$ and $\sigma_t = t$, yielding straight-line interpolation paths with the simplified target $\mathbf{v} = \boldsymbol{\epsilon} - \mathbf{x}_0$. This formulation forms the backbone of recent image editing models~\cite{flux2,wu2025qwenimage}. Our framework builds upon the flow-based MM-DiT, \ie Qwen-Image-Edit~\cite{wu2025qwenimage}.

\vspace{-0.5em}
\paragraph{\textbf{Text-centric Image Editing.}} 
Given a source image \( \mathbf{I}_{src} \) and a user-provided editing instruction \( \mathbf{p} \), the goal of instruction-based image editing is to generate a target image \( \mathbf{I}_{tgt} \) that semantically aligns with \( \mathbf{p} \) while preserving the non-target regions of \( \mathbf{I}_{src} \).
Distinct from general object or style editing, text-centric editing requires the model to manipulate textual elements--such as modifying, translating, deleting, or rearranging text--within an image. The generated text must accurately reflect the intent of the instruction, be correctly spelled, and maintain a consistent style with the original image, while strictly preserving the integrity of the background. In this work, we focus on this practically significant dimension of image editing tasks that has not yet been fully explored.

\begin{figure}[t]
  \centering
  \vspace{-1.5em}
  \includegraphics[width=1.0\linewidth]{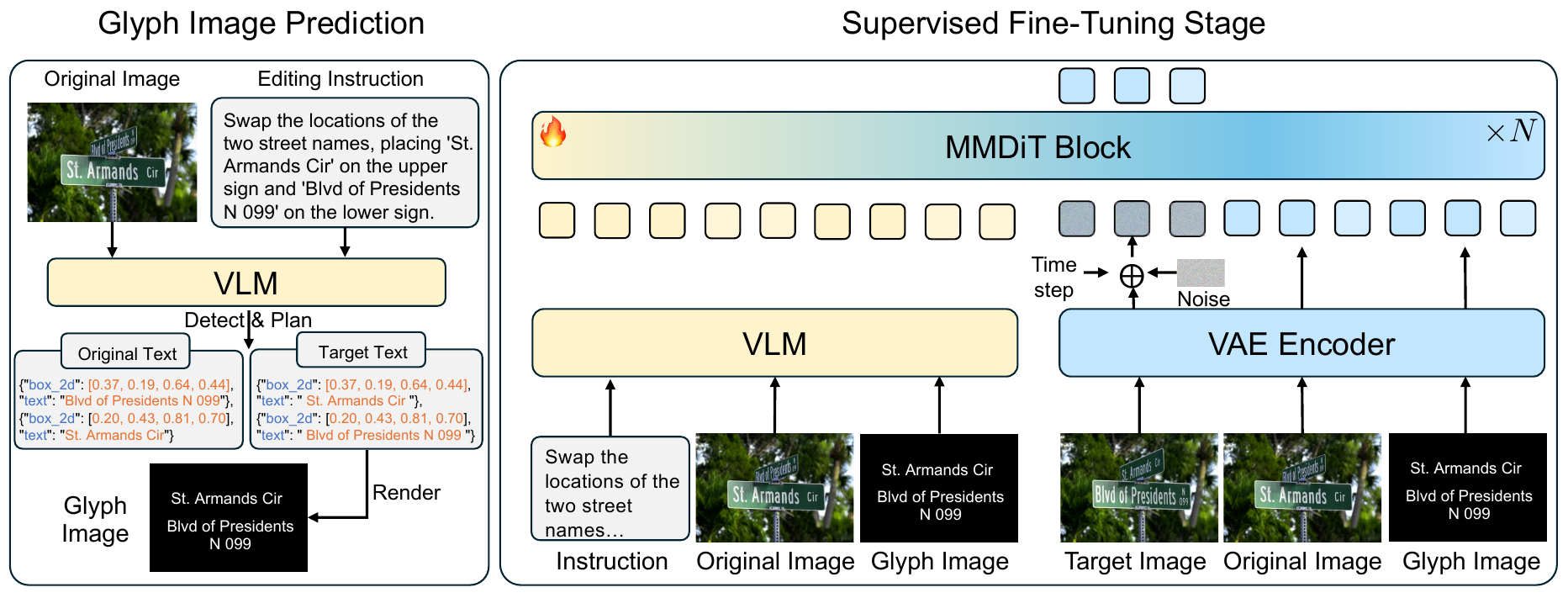}
  \caption{Overview of the glyph-guided supervised fine-tuning stage. A VLM first predicts the content and layout of the target text to render a glyph image. The original image, instruction, and glyph image are then jointly processed by the MM-DiT block to generate the target image.}
  \label{fig:sft}
  \vspace{-1.5em}
\end{figure}

\vspace{-0.75em}
\subsection{Supervised Fine-Tuning (SFT)}
\label{sec:sft}
\vspace{-0.5em}
While recent diffusion models have demonstrated impressive general editing capabilities, they often struggle with complex text-centric editing scenarios such as long-text replacement or multilingual translation, frequently generating hallucinated or misplaced characters.
To address this, we introduce a glyph-guided supervised fine-tuning stage to adapt a pre-trained image editing diffusion model for accurate text-centric editing. As illustrated in \cref{fig:sft}, our SFT pipeline comprises two key components: a glyph image prediction mechanism and a LoRA-based glyph-conditioned fine-tuning procedure.

\vspace{-0.75em}
\paragraph{\textbf{Glyph Image Prediction.}}
Given a source image $\mathbf{I}_{src}$ and an editing instruction $\mathbf{p}$, we employ Qwen3-VL-235B-A22B-Instruct~\cite{bai2025qwen3vl} to perform a two-step \textit{detect-and-plan} procedure. In the detection step, the VLM identifies all text regions of interest specified by the editing instruction in the source image and outputs their bounding boxes and content as structured tuples $\{(\mathbf{b}_i^{\text{orig}}, t_i^{\text{orig}})\}_{i=1}^{m}$, where $\mathbf{b}_i^{\text{orig}}$ denotes the normalized coordinates of the top-left and bottom-right corners, and $t_i^{\text{orig}}$ denotes the text string. In the planning step, the VLM interprets the editing instruction $\mathbf{p}$ and determines the target text content and spatial placement for each region, producing the target tuples $\{(\mathbf{b}_j^{\text{tgt}}, t_j^{\text{tgt}})\}_{j=1}^{n}$. 
Once the layout of target texts is determined, we render the glyph image $\mathbf{x}_{\text{glyph}}$ by drawing each target text string $t_j^{\text{tgt}}$ within its corresponding bounding box $\mathbf{b}_j^{\text{tgt}}$ on a blank canvas that matches the source image dimensions. 
To ensure high-quality and standardized glyph images, we use Python Pillow package~\cite{pillow} and the multilingual Arial~\cite{arial} font. For simplicity, all glyph images feature white text on a black background, providing a clear representation of the target texts. 
This glyph image explicitly includes the character content, spatial positions, and relative scales of all target texts, serving as a spatial prior for the subsequent editing process.

\vspace{-0.75em}
\paragraph{\textbf{LoRA-based Glyph-guided Fine-tuning.}}
We adopt a parameter-efficient fine-tuning strategy using Low-Rank Adaptation (LoRA)~\cite{hu2022lora} to train the editing model. The model takes three inputs: the editing instruction $\mathbf{p}$, the source image $\mathbf{I}_{src}$, and the glyph image $\mathbf{I}_{glyph}$. 
Specifically, the editing instruction, along with the original image and the glyph image, is first fed into the VLM to obtain semantically enriched text tokens. Meanwhile, $\mathbf{I}_{src}$ and $\mathbf{I}_{glyph}$ are encoded into latent representations via a VAE~\cite{kingma2013vae} and concatenated with the noisy latent $\mathbf{x}_t$ along the token dimension. We keep the original weights of the MM-DiT blocks frozen, and introduce LoRA modules only into the linear layers of the multi-modal attention mechanisms. This approach enables the model to effectively incorporate information from the glyph image, maintaining low training overhead.

\begin{figure}[t]
  \centering
  \includegraphics[width=1.0\linewidth]{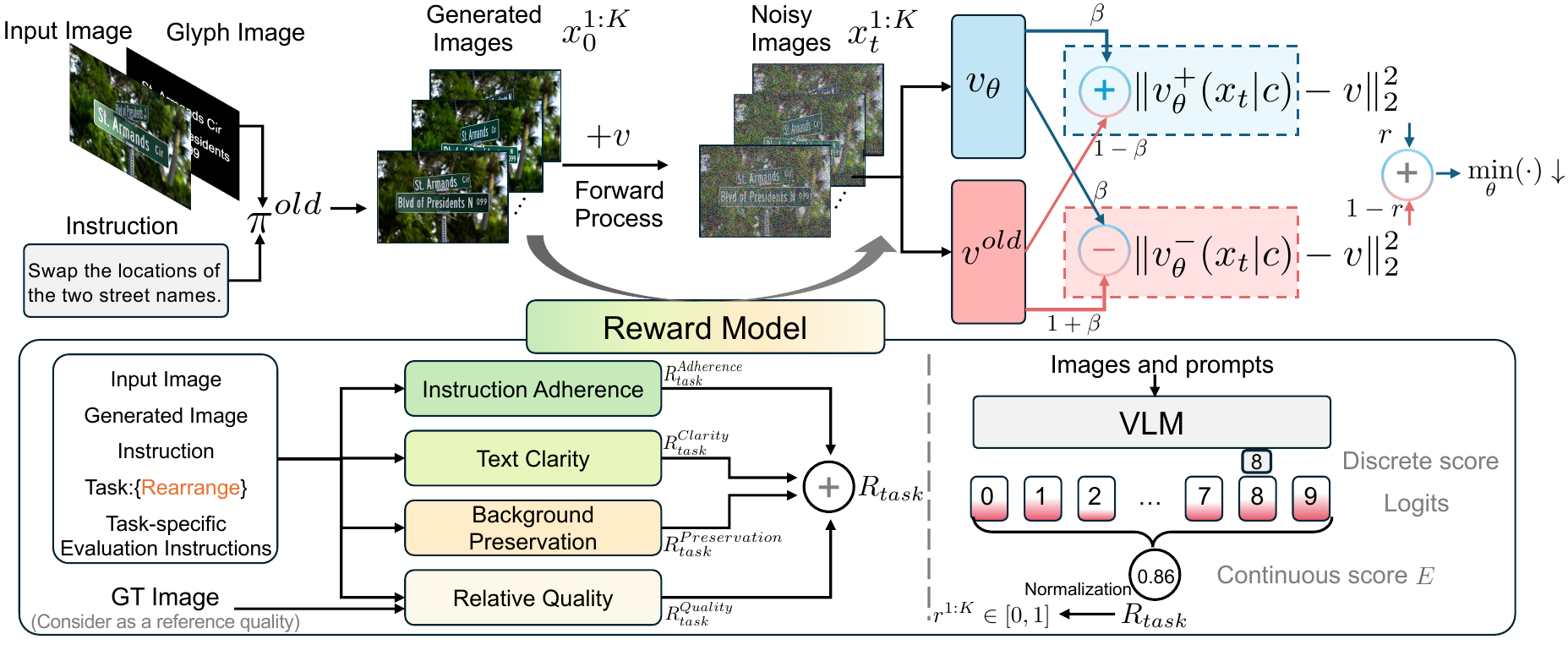}
  \caption{\textbf{Overview of the RL stage.} The model generates multiple candidate images, which are evaluated by four separate reward models targeting four dimensions. Each reward model leverages a Vision-Language Model to produce logit distributions over discrete scores, which are then converted to continuous expected values.}
  \label{fig:rl}
  \vspace{-1.2em}
\end{figure}

\vspace{-0.75em}
\subsection{Reinforcement Learning (RL)}
\label{sec:rlhf}
\vspace{-0.55em}
While the glyph-guided SFT stage equips the model with fundamental text editing capabilities, pixel-level losses cannot capture higher-level perceptual qualities such as legibility, instruction fidelity, and background coherence. 
To address this limitation, we introduce an RL stage that optimizes the diffusion model against a composite reward function tailored to the unique demands of text-centric image editing, as shown in \cref{fig:rl}. 

\vspace{-0.75em}
\paragraph{\textbf{DiffusionNFT-based Policy Optimization.}}
We adopt DiffusionNFT~\cite{zheng2025diffusionnft} as our policy optimization framework, which performs online reinforcement learning directly on the forward diffusion process via the flow matching objective. For each conditioning input $\mathbf{c}$ (comprising $\mathbf{I}_{src}$, $\mathbf{I}_{glyph}$, and instruction $\mathbf{p}$), we sample $K$ candidate images $\{\mathbf{x}_0^{1:K}\}$ from the current policy $\pi^{\text{old}}$ using an efficient ODE solver (\eg, DPM-Solver~\cite{lu2022dpmsolver}). Each candidate is then scored by the reward model to yield an optimality probability $r \in [0,1]$, which serves as a soft indicator of sample quality. The training objective jointly optimizes implicitly parameterized positive and negative policies:
{
\setlength{\abovedisplayskip}{2pt}
\setlength{\belowdisplayskip}{3pt}
\begin{small}
\begin{equation}
\mathcal{L}_{\text{RL}}(\theta) = \mathbb{E}_{\mathbf{c},\, \pi^{\text{old}}(\mathbf{x}_0|\mathbf{c}),\, t} \left[ r \left\| \mathbf{v}_\theta^+(\mathbf{x}_t, \mathbf{c}, t) - \mathbf{v} \right\|_2^2 + (1-r) \left\| \mathbf{v}_\theta^-(\mathbf{x}_t, \mathbf{c}, t) - \mathbf{v} \right\|_2^2 \right],
\end{equation}
\end{small}
where $\mathbf{v}$ denotes the target velocity, and the implicit positive and negative policies are:
\begin{align}
\mathbf{v}_\theta^+(\mathbf{x}_t, \mathbf{c}, t) &:= (1-\beta)\, \mathbf{v}^{\text{old}}(\mathbf{x}_t, \mathbf{c}, t) + \beta\, \mathbf{v}_\theta(\mathbf{x}_t, \mathbf{c}, t), \\
\mathbf{v}_\theta^-(\mathbf{x}_t, \mathbf{c}, t) &:= (1+\beta)\, \mathbf{v}^{\text{old}}(\mathbf{x}_t, \mathbf{c}, t) - \beta\, \mathbf{v}_\theta(\mathbf{x}_t, \mathbf{c}, t).
\end{align}
This formulation establishes a contrastive improvement direction: the positive branch pulls the $\mathbf{v}_\theta$ toward high-reward generations, while the negative branch pushes it away from low-reward ones, with the hyperparameter $\beta$ controlling the guidance strength.

\vspace{-0.75em}
\paragraph{\textbf{Task-Specific Multi-Dimensional Reward.}}
We design a reward function specifically tailored to the multi-faceted requirements of text-centric image editing. Rather than relying on a single holistic quality score, we decompose the reward into four complementary dimensions, each targeting a critical and distinct aspect of editing quality:

\noindent\textbf{(1) Instruction Adherence} ($R^{\text{Adherence}}_{\text{task}}$) evaluates whether the edited image faithfully executes the specified editing operation (\eg whether the correct text has been replaced, translated, or rearranged at the instructed locations). We prompt Qwen3-VL-235B-A22B-Instruct~\cite{bai2025qwen3vl} as a reward model (same below), which takes the original image, the edited image, and the editing instruction as input, and assesses the degree of semantic alignment between the instruction intent and the actual edit.

\noindent\textbf{(2) Text Clarity} ($R^{\text{Clarity}}_{\text{task}}$) measures the legibility and typographic quality of the rendered text. The VLM evaluates whether the generated characters are crisp, correctly spelled, and free from common artifacts such as blurriness, stroke distortion, or character merging---issues that are prevalent in text-centric editing tasks involving fine-grained glyph details.

\noindent\textbf{(3) Background Preservation} ($R^{\text{Preservation}}_{\text{task}}$) assesses whether non-target regions of the image remain unaltered after editing. Unintended modifications---such as color shifts, structural distortions, or texture degradation---are penalized to ensure that edits are strictly confined to the designated text regions.

\noindent\textbf{(4) Relative Quality} ($R^{\text{Quality}}_{\text{task}}$) compares the overall visual quality of the edited image against a \emph{reference image}, which can be either a ground-truth image or an output produced by a leading editing model. By grounding the evaluation in an explicit comparison target, this metric provides the VLM with a sharper perception of what constitutes a high-quality edit result, effectively establishing a quality anchor. As a result, the editing model is encouraged not only to produce outputs that merely appear satisfactory, but to generate edits that truly match or surpass the reference.

\vspace{-1.0em}
\paragraph{\textbf{Logit-Weighted Continuous Scoring.}}
Inspired by UniWorld-V2~\cite{li2025uniworldv2}, we adopt a logit-based continuous scoring mechanism to avoid the sparsity in single-integer reward signals. For each evaluation dimension, the VLM~\cite{bai2025qwen3vl} receives an input tuple $\mathbf{X} =(\mathbf{I}_{\text{src}}, \mathbf{I}_{\text{edited}}, T_{\text{prompt}})$, where $T_{\text{prompt}}$ is a dimension- and task-specific prompt instructing the VLM to rate the edit on a scale from 0 to 9. Instead of sampling a single token, we compute a softmax over the score token set $\mathcal{S} = \{0, 1, \ldots, 9\}$ at the designated decoding position and obtain the per-dimension reward as the normalized 
expected score:
{
\setlength{\abovedisplayskip}{2pt}
\setlength{\belowdisplayskip}{3pt}
\begin{equation}
    R^{\text{dim}}_{\text{task}}(\mathbf{X}) 
    \;=\; \frac{1}{\max(\mathcal{S})} \sum_{s \in \mathcal{S}} s \cdot 
    \frac{\exp(z_s)}{\sum_{s' \in \mathcal{S}} \exp(z_{s'})},
\label{eq:logit_reward}
\end{equation}
}
where $z_s$ is the logit for score token $s$. 
This soft scoring captures the VLM's full confidence distribution, yielding a smoother reward score.
The composite reward for each candidate is a weighted sum of all four dimensions:
{
\setlength{\abovedisplayskip}{2pt}
\setlength{\belowdisplayskip}{3pt}
\begin{equation}
R_{\text{task}} = \lambda_{\text{acc}}\, R^{\text{Adherence}}_{\text{task}} 
+ \lambda_{\text{cla}}\, R^{\text{Clarity}}_{\text{task}} 
+ \lambda_{\text{pre}}\, R^{\text{Preservation}}_{\text{task}} 
+ \lambda_{\text{qua}}\, R^{\text{Quality}}_{\text{task}},
\label{eq:composite_reward}
\end{equation}
}
where $\lambda_{\text{acc}}, \lambda_{\text{cla}}, \lambda_{\text{pre}}, 
\lambda_{\text{qua}}$ balance the relative importance of each dimension. Notably, the 
evaluation prompt for each dimension is further customized by editing task type 
(\eg, replace, translate, rearrange, or delete), enabling task-aware 
scoring criteria.

To feed $R_{\text{task}}$ into the policy optimization process, we convert it to an optimality probability 
$r \in [0,1]$ via intra-group normalization:
\begin{equation}
r(\mathbf{x}_0, \mathbf{c}) = \frac{1}{2} + \frac{1}{2}\, \text{clip}\!\left[
\frac{R_{\text{task}}(\mathbf{x}_0, \mathbf{c}) 
- \mu_{\mathbf{c}}}{\sigma_{\mathbf{c}}},\; -1,\; 1\right],
\label{eq:optimality}
\end{equation}
where $\mu_{\mathbf{c}}$ and $\sigma_{\mathbf{c}}$ are the mean and standard deviation of $R_{\text{task}}$ over the $K$ candidates sampled for conditioning input $\mathbf{c}$. 
This normalization ensures that $r$ reflects relative quality within each sample group.
The value of $r$ is then used to directly weight the positive and negative branches in $\mathcal{L}_{\text{RL}}$.

\vspace{-1.0em}
\section{Dataset and Benchmark}
\label{sec:dataset_and_benchmark}

In this section, we first introduce our data construction pipeline, which generates high-quality source–target image pairs via two complementary routes (Sec.~\ref{sec:dataset}). We then introduce our evaluation benchmark, covering bilingual (Chinese--English) and multilingual (15 languages) text-centric editing scenarios (Sec.~\ref{sec:benchmark}).

\begin{figure}[t]
  \centering
  \vspace{-1.5em}
  \includegraphics[width=0.85\linewidth]{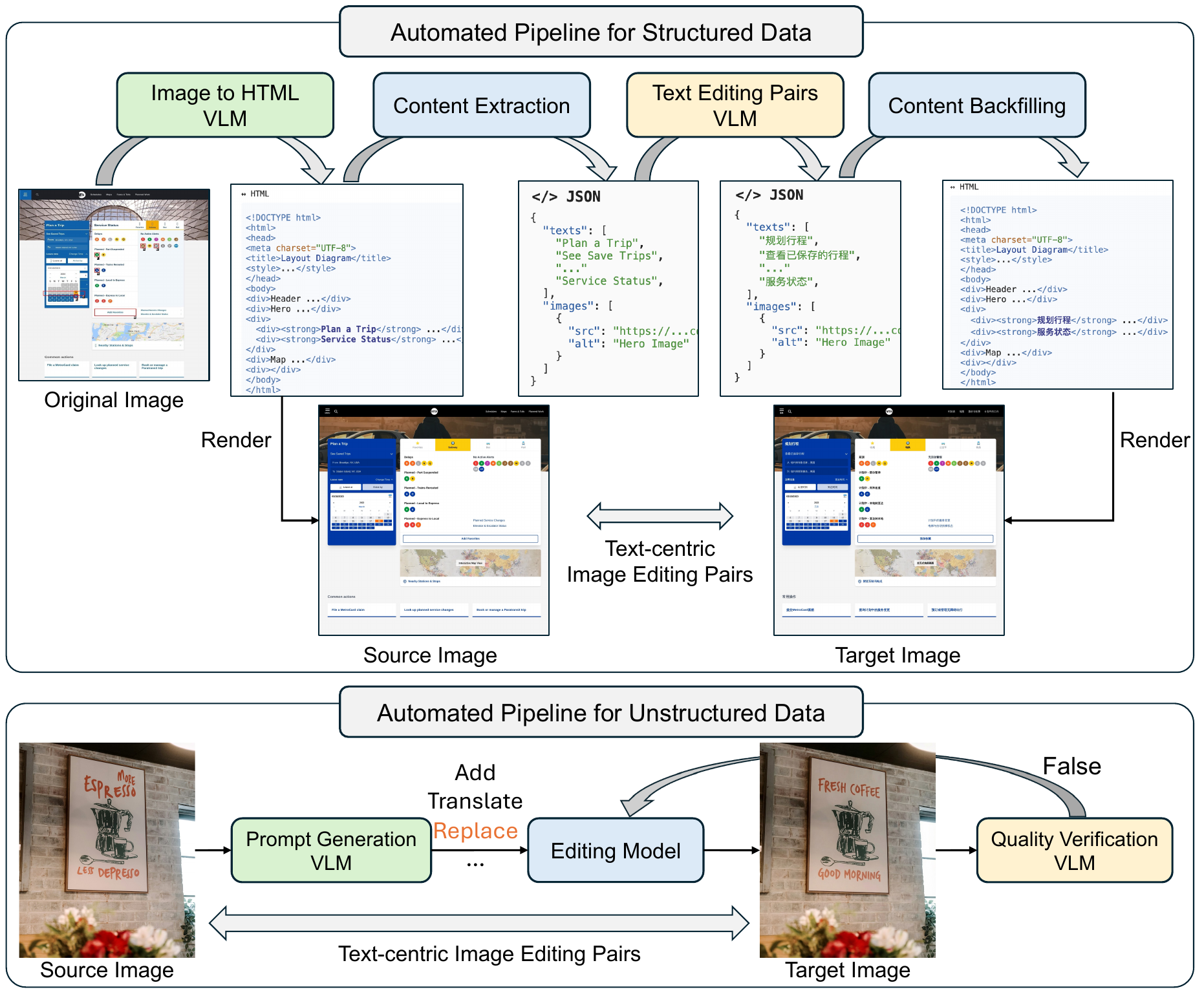}
  \caption{Overview of our data construction pipelines. \textbf{Top:} the \emph{structured} pipeline converts a source image to HTML, extracts and edits text content via a VLM, and renders both source and target images through a headless browser, yielding pixel-perfect editing pairs. \textbf{Bottom:} the \emph{unstructured} pipeline uses a VLM to propose editing instructions, executes edits with a generative model, and iteratively verifies quality until all acceptance criteria are met.}
  \label{fig:data_pipeline}
  \vspace{-1.5em}
\end{figure}

\vspace{-1.0em}
\subsection{Dataset}
\label{sec:dataset}

Our dataset is designed to cover a broad spectrum of text-centric editing tasks. We define 7 operation types: \emph{Add} (inserting new text), \emph{Replace} (substituting existing text with different content), \emph{Delete} (removing text regions), \emph{Rearrange} (permuting the spatial layout of text elements), \emph{Translate} (converting text between languages), \emph{Change Style} (modifying font, color, or typographic attributes), and \emph{Combined} (composing multiple atomic operations in a single instruction).
Based on the nature of the source images, we identified two fundamentally different data categories—structured and unstructured—and designed a dedicated construction process for each, as illustrated in \cref{fig:data_pipeline}.

\vspace{-0.5em}
\paragraph{\textbf{Structured Data.}}
For images with well-organized layouts and relatively uniform text styles---such as web page screenshots, mobile app interfaces, presentation slides, figures, tables, documents, and infographics---we adopt a novel HTML-based construction pipeline (\cref{fig:data_pipeline}, top). We collected a wide range of source images from various open-source datasets, including Leopard~\cite{jia2024leopard}, The Cauldron~\cite{laurenccon2024matters}, WebSight~\cite{laurenccon2024websight}, \etc. The pipeline proceeds in four stages:

\noindent\textbf{(1) Image to HTML.}
Given an original image $\mathbf{I}_{\text{ori}}$, we employ a VLM~\cite{google2025gemini-3.0-pro} to convert it into an HTML representation $\mathcal{H}_{\text{src}}$ that preserves both the visual
layout and the textual content as faithfully as possible, using Tailwind~\cite{tailwind} CSS for styling.

\noindent\textbf{(2) Content Extraction.}
We use BeautifulSoup4~\cite{beautifulsoup4} to parse
$\mathcal{H}_{\text{src}}$ and extract a structured JSON record containing
all text elements and image elements.  For images, we validate external URLs
and retrieve semantically similar replacements from the web when the
original links are broken, ensuring rendering fidelity.

\noindent\textbf{(3) Text Editing Pairs.}
Given the $N$ text entries extracted from $\mathcal{H}_{\text{src}}$, we
employ a lightweight VLM Qwen3-30B-A3B~\cite{bai2025qwen3vl} to generate the corresponding edited
texts for each operation type. Specifically, for \emph{translation}, all $N$ entries are
translated into the target language; for \emph{replacement}, 1 to
$N$ entries are randomly selected, and the VLM produces semantically
plausible substitutes.

\noindent\textbf{(4) Content Backfilling.}
Once the target texts are determined, we replace the original text entries in $\mathcal{H}{\text{src}}$ at their exact positions while keeping all other HTML elements unchanged, yielding the modified HTML code $\mathcal{H}{\text{tgt}}$.

\noindent\textbf{(5) Rendering.}
Both $\mathcal{H}_{\text{src}}$ and $\mathcal{H}_{\text{tgt}}$ are rendered
into images via Playwright~\cite{playwright}.
As the rendering process is fully deterministic and
$\mathcal{H}_{\text{src}}$ and $\mathcal{H}_{\text{tgt}}$ differ only in
the designated text entries, the resulting source image
$\mathbf{I}_{\text{src}}$ and target image $\mathbf{I}_{\text{tgt}}$ are
guaranteed to be pixel-perfect identical in all non-target regions.  

Another key advantage of our HTML-based pipeline is its natural extensibility to multilingual data.  Before constructing editing pairs, we can first translate all text entries in $\mathcal{H}_{\text{src}}$ into a target language to obtain a multilingual-based HTML, and then apply the same
pair-construction workflow on top of it.  In this work, we cover 15 widely used languages: English, Chinese, Hindi, Spanish, French, Arabic, Portuguese, Bengali, Russian, German, Korean, Japanese, Thai, Indonesian, and Vietnamese.
This automated pipeline enables efficient construction of large-scale, high-quality, and linguistically diverse text-centric editing pairs.

\vspace{-0.5em}
\paragraph{\textbf{Unstructured Data.}}
For images with complex layouts, diverse typography, and text is tightly entangled with complex visual backgrounds~(\eg, street signs, product packaging, posters, and scene photographs), HTML-based reconstruction cannot faithfully recover the intricate visual styles. 
To address this, we develop an automated edit-verify-and-retry pipeline that operates directly at the image level (\cref{fig:data_pipeline}, bottom).
Given a source image $\mathbf{I}_{\text{src}}$, a VLM~\cite{google2025gemini-3.0-pro} first analyzes its
content and proposes a set of plausible editing instructions covering various task types.
Each instruction is then executed by an editing model~\cite{google2025gemini-3.0-pro-image,flux2,wu2025qwenimage} to produce a candidate
target image $\hat{\mathbf{I}}_{\text{tgt}}$.
A separate VLM-based~\cite{google2025gemini-3.0-pro} verifier evaluates the candidate against the source
image and instruction on instruction adherence, text legibility, and
background preservation. Failed candidates are fed back with verification feedback for re-execution; only candidates passing all checks are retained as valid training pairs.
This closed-loop workflow ensures high-quality data for visually challenging scenes, filters out ambiguous or infeasible edits, and provides a clean, diverse complement to the structured data.

\paragraph{\textbf{Dataset Statistics.}} 
We constructed a training set containing 330K samples, including approximately 160K unstructured text pairs and 170K structured text pairs.
The dataset spans 7 editing operation types and covers 15 languages. The number of editing regions per sample ranges from 1 to over 13, and the edited text length varies widely from under 20 characters to over 1{,}000 characters, demonstrating the diversity and complexity of our dataset. Detailed statistical breakdowns are provided in the supplementary material.

\begin{figure}[t] %
     \centering
     \vspace{-1.5em}
     \includegraphics[width=1.0\textwidth]{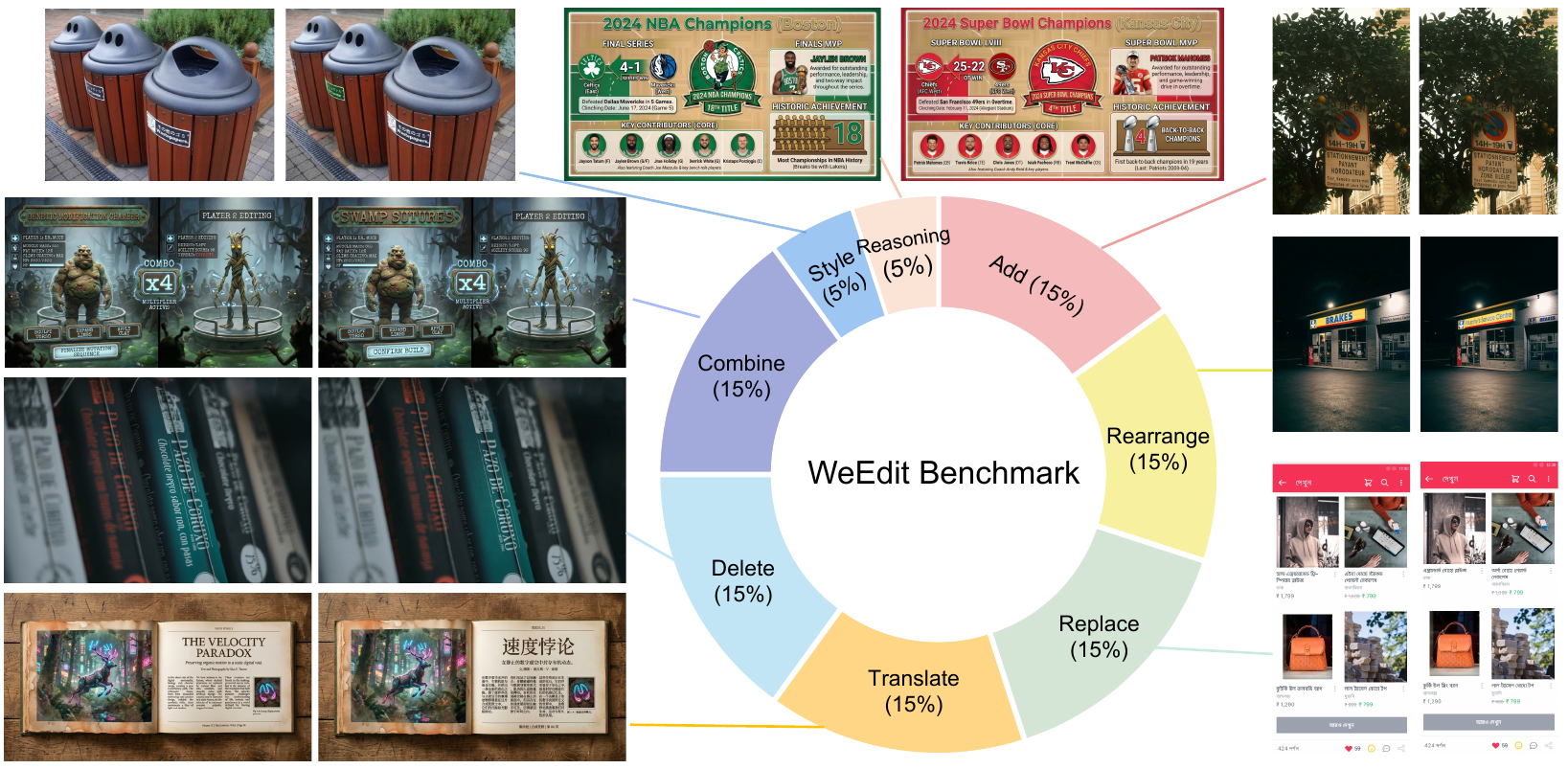}
    \caption{Overview of the proposed WeEdit Benchmark.}
    \label{fig:benchmark}
    \vspace{-1.5em}
\end{figure}

\vspace{-1.5em}
\subsection{Benchmark}
\label{sec:benchmark}

As illustrated in \cref{fig:benchmark}, we construct a comprehensive benchmark that covers a diverse set of text-centric editing operations, supports multiple languages, and evaluates model performance along three complementary dimensions.

\vspace{-0.5em}
\paragraph{\textbf{Diverse Editing Operations.}} 
Our benchmark encompasses eight distinct task categories to thoroughly assess model capabilities. Six of these focus on direct textual content manipulation: \textit{Add}, \textit{Replace}, \textit{Delete}, \textit{Rearrange}, \textit{Translate}, and \textit{Combined} (executing multiple atomic operations simultaneously). Furthermore, we include \textit{Change Style} to evaluate the modification of typographic attributes (\eg, font, color). In addition, we introduce a more advanced \textit{Reasoning} task that goes beyond the operation types found in the training set. This task requires the model to first deduce the correct target text based on knowledge or logical context before performing the edit. For instance, replacing ``2024 NBA Champions (Boston)'' with ``2024 Super Bowl Champions'' requires the model to recall relevant factual knowledge (\eg, team names, scores, and player details) to obtain the target text content before editing.

\vspace{-0.5em}
\paragraph{\textbf{Comprehensive Language Coverage.}} 
To evaluate the robustness of editing models across different linguistic contexts, the benchmark is provided in two distinct versions, each containing 2,000 test cases. The \textit{Bilingual Benchmark} focuses on Chinese and English, representing typical high-frequency scenarios. Meanwhile, the \textit{Multilingual Benchmark} extends the evaluation to 15 widely used languages, testing the models' generalization ability to diverse character sets and complex glyph structures.

\paragraph{\textbf{Evaluation Metrics.}} 
We adopt a VLM-based automatic evaluation protocol, employing Gemini-3-Pro~\cite{google2025gemini-3.0-pro} as an impartial judge to score generated images across three core dimensions: \textit{Instruction Adherence}~(IA), which measures the degree to which the edited image faithfully fulfills each requirement specified in the editing instruction; \textit{Text Clarity}~(TC), which assesses the legibility and spelling correctness of the generated text; and \textit{Background Preservation}~(BP), which quantifies the visual integrity of non-target regions. The judge is prompted to first generate a detailed chain-of-thought~\cite{wei2022cot} rationale that examines the edited image against the evaluation criterion, and then assign a score on a 0--9 scale, thereby producing fine-grained quality assessments.

\vspace{-0.5em}
\section{Experiment}
\label{sec:experiment}

\vspace{-0.5em}
\subsection{Experimental Setup}

\vspace{-0.5em}
\paragraph{\textbf{Dataset and Evaluation Metrics.}}
We train our models on the proposed WeEdit dataset. For evaluation, we utilize the WeEdit Benchmark, consisting of a Bilingual Benchmark (Chinese and English) and a Multilingual Benchmark, each containing 2,000 meticulously curated test cases to thoroughly assess model capabilities. Following ~\cref{sec:benchmark}, we employ Gemini-3-Pro~\cite{google2025gemini-3.0-pro} to assess the edited images from three dimensions.

\vspace{-0.5em}
\paragraph{\textbf{Implementation Details.}}
During the SFT stage, we fine-tune Qwen-Image-Edit-2509~\cite{qwen2025Qwen-Image-Edit-2509} using LoRA with a rank of 256. The model is optimized with AdamW at a learning rate of 5e-5, trained for 8,000 steps.
In the RL stage, we initialize the model from the SFT weights and further fine-tune it with LoRA of rank 256 and AdamW at a learning rate of $5{\times}10^{-5}$ for 140 epochs. The reward weights are set to $\lambda_\text{acc}{=}0.3$, $\lambda_\text{cla}{=}0.3$, $\lambda_\text{pre}{=}0.15$, and $\lambda_\text{qua}{=}0.25$.

\vspace{-0.5em}
\subsection{Main Results}

\vspace{-0.5em}
\paragraph{\textbf{Baseline Methods.}}
To evaluate WeEdit, we compare it against 15 previous SOTA baselines, including 4 prominent proprietary models~\cite{google2025gemini-3.0-pro-image,google2025gemini-2.5-flash-image,openai2025gpt-image-1.5,bytedance2025seedream4.5} and 11 leading open-source models~\cite{wu2025omnigen2,deng2025bagel,cui2025emu3.5,li2025uniworldv2,flux2,qwen2025Qwen-Image-Edit-2509,qwen2025Qwen-Image-Edit-2511,team2025longcat,liu2025step1x,cao2025hunyuanimage,team2026firered}. The evaluation is conducted on both our Bilingual and Multilingual benchmarks, systematically covering 8 diverse editing operations.  For each operation, the edited images are rigorously assessed across three complementary criteria: Instruction Adherence (IA), Text Clarity (TC), and Background Preservation (BP).

\begin{table}[t]
\caption{\textbf{Quantitative Results on the Bilingual Benchmark.} We compare WeEdit with \colorbox{close_source}{proprietary} and \colorbox{open_source}{open-source} models across 8 editing operations. IA, TC, and BP denote Instruction Adherence, Text Clarity, and Background Preservation. The best open-source results are in \textbf{bold}, with \colorbox{top}{top-3} highlighted. Our method significantly improves text editing over the \colorbox{baseline}{base model}, outperforms all open-source models, and surpasses most proprietary systems.}
\centering
\tabcolsep=0.02cm
\ra{1.1}
\scalebox{0.39}{
\begin{tabular}{@{}lccc|ccc|ccc|ccc|ccc|ccc|ccc|ccc|ccc@{}}
\toprule                           
\multirow{2}{*}{\textbf{Model}}            & \multicolumn{3}{c|}{\textbf{Add}}          & \multicolumn{3}{c|}{\textbf{Replace}}      & \multicolumn{3}{c|}{\textbf{Delete}}       & \multicolumn{3}{c|}{\textbf{Rearrange}}    & \multicolumn{3}{c|}{\textbf{Translate}}    & \multicolumn{3}{c|}{\textbf{Style}} & \multicolumn{3}{c|}{\textbf{Combined}}     & \multicolumn{3}{c|}{\textbf{Reasoning}}    & \multicolumn{3}{c}{\textbf{Overall}}      \\ \cmidrule(l){2-28}
                                  & IA & TC & BP & IA & TC & BP & IA & TC & BP & IA & TC & BP & IA & TC & BP & IA & TC & BP & IA & TC & BP & IA & TC & BP & IA & TC & BP \\ \midrule
\cellcolor{close_source}{Gemini-3-Pro-Image~\cite{google2025gemini-3.0-pro-image}}                      & \textbf{9.38}                   & \textbf{9.66}                   & \textbf{9.43}                   & \textbf{9.06}                   & \textbf{8.80}                    & \textbf{8.05}                   & \textbf{8.33}                   & \textbf{8.56}                   & \textbf{8.04}                   & \textbf{8.05}                   & \textbf{8.49}                   & \textbf{8.26}                   & \textbf{8.08}                   & \textbf{9.07}                   & \textbf{9.55}                   & \textbf{9.76}                   & 9.51                   & \textbf{9.50}                    & \textbf{9.31}                   & \textbf{9.72}                   & \textbf{9.45}                   & \textbf{4.91}                   & \textbf{9.47}                   & \textbf{9.42}                   & \textbf{8.58}                   & \textbf{9.10}                    & \textbf{8.85}                   \\
\cellcolor{close_source}{Gemini-2.5-Flash-Image~\cite{google2025gemini-2.5-flash-image}}                    & 5.88                   & 8.06                   & 7.58                   & 3.35                   & 6.05                   & 7.50                    & 6.01                   & 8.17                   & 6.05                   & 1.77                   & 6.86                   & 7.11                   & 1.75                   & 6.06                   & 8.83                   & 7.95                   & \textbf{9.54}                   & 9.58                   & 4.08                   & 7.57                   & 8.83                   & 1.70                    & 4.82                   & 8.76                   & 3.92                   & 7.14                   & 7.80                    \\
\cellcolor{close_source}{GPT-Image-1.5~\cite{openai2025gpt-image-1.5}}                     & 8.02                   & 9.32                   & 6.80                    & 7.16                   & 7.58                   & 5.55                   & 6.43                   & 6.49                   & 3.82                   & 5.44                   & 6.91                   & 5.35                   & 4.09                   & 6.97                   & 7.71                   & 9.30                    & 9.44                   & 8.31                   & 8.12                   & 9.12                   & 6.34                   & 3.37                   & 7.02                   & 7.93                   & 6.52                   & 7.78                   & 6.15                   \\
\cellcolor{close_source}{Seedream4.5~\cite{bytedance2025seedream4.5}}                       & 6.86                   & 8.21                   & 6.40                    & 7.23                   & 7.93                   & 5.91                   & 6.60                    & 7.37                   & 4.70                    & 5.29                   & 7.08                   & 5.36                   & 4.97                   & 6.93                   & 7.15                   & 8.76                   & 9.08                   & 8.37                   & 7.33                   & 8.62                   & 7.66                   & 2.17                   & 5.69                   & 7.67                   & 6.29                   & 7.66                   & 6.38                   \\ \midrule
\cellcolor{open_source}{OmniGen2~\cite{wu2025omnigen2}}                          & 1.72                   & 3.11                   & 3.86                   & 1.26                   & 3.30                    & 3.61                   & 1.53                   & 3.54                   & 2.52                   & 1.07                   & 3.34                   & 3.32                   & 1.00                      & 3.35                   & 3.85                   & 3.67                   & 5.29                   & 4.21                   & 1.21                   & 3.41                   & 3.92                   & 1.01                   & 4.15                   & 6.24                   & 1.40                    & 3.48                   & 3.68                   \\
\cellcolor{open_source}{BAGEL~\cite{deng2025bagel}}                             & 2.32                   & 3.03                   & 5.99                   & 1.48                   & 2.41                   & 4.30                    & 3.03                   & 6.46                   & 2.92                   & 1.07                   & 3.53                   & 5.66                   & 1.03                   & \cellcolor{top}{4.88}  & 7.68                   & 7.36                   & 7.76                   & 8.98                   & 1.44                   & 2.33                   & 5.83                   & 1.02                   & 4.42                   & 8.89                   & 1.97                   & 4.01                   & 5.75                   \\
\cellcolor{open_source}{Emu3.5~\cite{cui2025emu3.5}}                            & 4.89                   & 6.60                    & 3.72                   & 3.26                   & 4.60                    & 2.81                   & 4.76                   & 5.20                    & 2.29                   & 1.46                   & 3.94                   & 3.36                   & 1.27                   & 2.73                   & 5.93                   & 8.27                   & 8.66                   & 8.41                   & 3.96                   & 5.98                   & 4.13                   & 1.23                   & 3.34                   & 6.32                   & 3.42                   & 4.96                   & 4.07                   \\
\cellcolor{open_source}{UniWorld-v2~\cite{li2025uniworldv2}}                       & 5.15                   & 7.16                   & 5.41                   & 3.27                   & 5.06                   & 4.29                   & 4.23                   & 7.13                   & 2.88                   & 1.47                   & 4.33                   & 4.39                   & 1.32                   & 2.36                   & 8.50                    & 8.78                   & 8.98                   & 8.55                   & 3.52                   & 6.22                   & 5.08                   & 1.22                   & 4.10                    & 7.97                   & 3.34                   & 5.49                   & 5.41                   \\
\cellcolor{open_source}{FLUX.2-dev~\cite{flux2}}                         & 5.40                    & \cellcolor{top}{7.59}  & 4.92                   & 3.59                   & 6.09                   & 4.95                   & 3.07                   & 8.14                   & 5.27                   & 1.44                   & \cellcolor{top}{5.73}  & 5.84                   & 1.24                   & 3.59                   & \cellcolor{top}{8.78}  & 8.04                   & 8.88                   & 8.31                   & 4.60                    & \cellcolor{top}{7.31}  & 5.52                   & \cellcolor{top}{1.26}  & \cellcolor{top}{\textbf{6.33}} & \cellcolor{top}{\textbf{9.55}} & 3.37                   & \cellcolor{top}{6.53}  & 6.19                   \\
\cellcolor{open_source}{Qwen-Image-Edit-2511~\cite{qwen2025Qwen-Image-Edit-2511}}              & 4.80                    & 5.94                   & 3.73                   & 3.36                   & 3.73                   & 3.77                   & 3.89                   & 4.46                   & 3.34                   & 1.42                   & 2.67                   & 3.95                   & 1.28                   & 1.60                    & 6.97                   & 8.50                    & 8.88                   & 8.48                   & 3.27                   & 4.21                   & 4.23                   & 1.09                   & 1.97                   & 6.04                   & 3.18                   & 3.93                   & 4.63                   \\
\cellcolor{open_source}{LongCat-Image-Edit~\cite{team2025longcat}}                & 5.23                   & 6.10                    & 6.80                    & 3.03                   & 4.51                   & 5.16                   & 4.65                   & 7.18                   & 5.84                   & 1.33                   & 4.91                   & 5.92                   & 1.29                   & 4.42                   & 8.29                   & 8.40                    & 8.95                   & 9.27  & 3.89                   & 5.47                   & 6.87                   & 1.17                   & 5.05                   & 8.38                   & 3.39                   & 5.59                   & 6.71                   \\
\cellcolor{open_source}{Step1X-Edit-v1.2~\cite{liu2025step1x}}                  & 2.73                   & 3.98                   & 7.38                   & 2.44                   & 3.22                   & \cellcolor{top}{6.75}  & 5.68                   & 7.01                   & \cellcolor{top}{6.45}  & 1.26                   & 3.46                   & 4.76                   & 1.34                   & 2.84                   & 7.64                   & 7.78                   & 8.54                   & 8.67                   & 2.12                   & 3.79                   & \cellcolor{top}{7.89}  & 1.09                   & \cellcolor{top}{5.81}  & \cellcolor{top}{9.37}  & 2.78                   & 4.36                   & \cellcolor{top}{7.03}  \\
\cellcolor{open_source}{HY-image-3-instruct~\cite{cao2025hunyuanimage}}         & \cellcolor{top}{6.04}                   & 7.55                   & \cellcolor{top}{7.68}  & 3.91                   & 5.42                   & 5.79                   & 6.32                   & 7.85                   & 4.69                   & 1.73                   & 4.26                   & \cellcolor{top}{6.59}  & \cellcolor{top}{1.45}  & 3.52                   & 8.67                   & \cellcolor{top}{\textbf{9.43}} & \cellcolor{top}{\textbf{9.35}} & \cellcolor{top}{9.52}  & 4.73                   & 6.62                   & 7.43                   & 1.12                   & 4.67                   & 8.65                   & \cellcolor{top}{4.16}  & 5.99                   & \cellcolor{top}{7.03}  \\
\cellcolor{open_source}{FireRed-Image-Edit~\cite{team2026firered}}                & 5.44                   & 7.54                   & 7.67                   & \cellcolor{top}{4.58}  & \cellcolor{top}{6.30}   & 6.26                   & \cellcolor{top}{6.80}   & \cellcolor{top}{8.81}  & 5.97                   & \cellcolor{top}{1.76}  & 4.78                   & 5.52                   & 1.39                   & 2.67                   & 8.44                   & \cellcolor{top}{9.30}   & \cellcolor{top}{9.23}  & \cellcolor{top}{\textbf{9.78}} & 4.19                   & 7.23                   & 7.48                   & 1.25                   & \cellcolor{top}{5.45}  & 8.90                    & 4.15                   & 6.33                   & \cellcolor{top}{7.14}  \\ \midrule
\cellcolor{baseline}{Qwen-Image-Edit-2509~\cite{qwen2025Qwen-Image-Edit-2509}} & 4.96                   & 7.05                   & 6.55                   & 3.44                   & 5.17                   & 5.90                    & 4.97                   & 7.84                   & 5.07                   & 1.35                   & 4.90                    & 5.91                   & 1.26                   & 2.72                   & \cellcolor{top}{8.77}  & 8.97                   & 9.15                   & 9.11                   & 3.90                    & 6.48                   & 7.24                   & 1.22                   & 5.17                   & 8.51                   & 3.49                   & 5.84                   & 6.80                    \\
 \textbf{WeEdit-SFT (ours)}                           & \cellcolor{top}{7.83}               & \cellcolor{top}{8.54}  & \cellcolor{top}{9.18}  & \cellcolor{top}{6.36}  & \cellcolor{top}{6.75}  & \cellcolor{top}{7.76}  & \cellcolor{top}{\textbf{9.28}} & \cellcolor{top}{9.12}  & \cellcolor{top}{\textbf{8.43}} & \cellcolor{top}{7.06}  & \cellcolor{top}{7.18}  & \cellcolor{top}{8.50}   & \cellcolor{top}{5.14}  & \cellcolor{top}{5.24}  & 8.72                   & 8.87                   & 8.54                   & 8.94                   & \cellcolor{top}{7.28}  & \cellcolor{top}{8.35}  & \cellcolor{top}{9.14}  & \cellcolor{top}{2.08}  & 2.48                   & 8.56                   & \cellcolor{top}{6.99}  & \cellcolor{top}{7.33}  & \cellcolor{top}{8.63}  \\
\textbf{WeEdit-RL (ours)}                           & \cellcolor{top}{\textbf{8.12}}                    & \cellcolor{top}{\textbf{8.99}} & \cellcolor{top}{\textbf{9.40}} & \cellcolor{top}{\textbf{6.90}} & \cellcolor{top}{\textbf{7.77}} & \cellcolor{top}{\textbf{8.63}} & \cellcolor{top}{8.92}   & \cellcolor{top}{\textbf{9.49}} & \cellcolor{top}{8.09}  & \cellcolor{top}{\textbf{7.82}} & \cellcolor{top}{\textbf{7.94}} & \cellcolor{top}{\textbf{9.00}}  & \cellcolor{top}{\textbf{6.36}} & \cellcolor{top}{\textbf{7.11}} & \cellcolor{top}{\textbf{9.44}} & \cellcolor{top}{9.41}  & \cellcolor{top}{9.24}  & \cellcolor{top}{9.47}                   & \cellcolor{top}{\textbf{7.50}} & \cellcolor{top}{\textbf{8.89}} & \cellcolor{top}{\textbf{9.38}} & \cellcolor{top}{\textbf{2.57}} & 5.24                   & \cellcolor{top}{8.99}     & \cellcolor{top}{\textbf{7.47}} & \cellcolor{top}{\textbf{8.19}} & \cellcolor{top}{\textbf{9.01}} \\
\textit{vs. \cellcolor{baseline}{Baseline}}                    &\textbf{\green{+3.16}}                   & \textbf{\green{+1.94}}                   & \textbf{\green{+2.85}}                   & \textbf{\green{+3.46}}                   & \textbf{\green{+2.6}}                   & \textbf{\green{+2.73}}                   & \textbf{\green{+3.95}}                   & \textbf{\green{+1.65}}                   & \textbf{\green{+3.02}}                   & \textbf{\green{+6.47}}                    & \textbf{\green{+3.04}}                   & \textbf{\green{+3.09}}                   & \textbf{\green{+5.10}}                   & \textbf{\green{+4.39}}                   & \textbf{\green{+0.67}}                   & \textbf{\green{+0.44}}                   & \textbf{\green{+0.09}}                   & \textbf{\green{+0.36}}                   & \textbf{\green{+3.6}}                   & \textbf{\green{+2.41}}                   & \textbf{\green{+2.14}}                   & \textbf{\green{+1.35}}                   & \textbf{\green{+0.07}}                   & \textbf{\green{+0.48}}                   & \textbf{\green{+3.98}}                   & \textbf{\green{+2.35}}                   & \textbf{\green{+2.21}}                   \\ \bottomrule
\end{tabular}}
\label{tab: Quantitative Comparison Bilingual}
\vspace{-1.0em}
\end{table}

\vspace{-0.5em}
\paragraph{\textbf{Quantitative Comparison.}}
\cref{tab: Quantitative Comparison Bilingual} and \cref{tab: Quantitative Comparison Multilingual} present results on the Bilingual and Multilingual benchmarks, respectively. We observe that text-centric image editing remains a fundamental weakness of existing proprietary and, in particular, open-source models.
On the Bilingual benchmark, the best-performing open-source baseline achieves only 4.16 overall IA, and the situation deteriorates further on more challenging tasks: on \emph{Rearrange} and \emph{Translate}, all open-source baselines score below 1.8 IA, indicating near-complete failure. Even strong proprietary systems (\eg GPT-Image-1.5 and Seedream4.5) exhibit quality degradation on these tasks, confirming that text-centric editing poses challenges qualitatively different from conventional object or style editing.

WeEdit establishes a new SOTA among open-source models by a clear margin and is surpassed only by the proprietary Gemini-3-Pro-Image. On the Bilingual benchmark, our RL version~(WeEdit-RL) achieves 7.47/8.19/9.01 in IA/TC/BP, improving over the base model by +3.98/+2.35/+2.21 and surpassing the previous best open-source results by a clear margin. 
The gains are more pronounced in tasks such as \emph{Translate} and \emph{Rearrange}.
Moreover, WeEdit demonstrates top-tier performance across all other editing operations.

\begin{table}[t]
\caption{Quantitative Results on the Multilingual Benchmark.}
\centering
\tabcolsep=0.02cm
\ra{1.1}
\scalebox{0.39}{
\begin{tabular}{@{}lccc|ccc|ccc|ccc|ccc|ccc|ccc|ccc|ccc@{}}
\toprule
\multirow{2}{*}{\textbf{Model}}            & \multicolumn{3}{c|}{\textbf{Add}}          & \multicolumn{3}{c|}{\textbf{Replace}}      & \multicolumn{3}{c|}{\textbf{Delete}}       & \multicolumn{3}{c|}{\textbf{Rearrange}}    & \multicolumn{3}{c|}{\textbf{Translate}}    & \multicolumn{3}{c|}{\textbf{Style}} & \multicolumn{3}{c|}{\textbf{Combined}}     & \multicolumn{3}{c|}{\textbf{Reasoning}}    & \multicolumn{3}{c}{\textbf{Overall}}      \\ \cmidrule(l){2-28}
                                  & IA & TC & BP & IA & TC & BP & IA & TC & BP & IA & TC & BP & IA & TC & BP & IA & TC & BP & IA & TC & BP & IA & TC & BP & IA & TC & BP \\ \midrule
\cellcolor{close_source}{Gemini-3-Pro-Image~\cite{google2025gemini-3.0-pro-image}}     & 9.59     & 9.35    & 8.83         & 8.39     & 8.21    & 6.97         & 8.17     & 8.08    & 6.71         & 7.37     & 7.97    & 7.11         & 6.34     & 7.60     & 8.66         & 8.82     & 8.77    & 7.78         & 8.83     & 8.77    & 8.12         & 6.04     & 8.87    & 9.29         & 8.05     & 8.38    & 7.82         \\
\cellcolor{close_source}{Gemini-2.5-Flash-Image~\cite{google2025gemini-2.5-flash-image}} & 6.23     & 8.12    & 8.49         & 3.25     & 6.42    & 7.56         & 6.31     & 8.06    & 6.32         & 1.84     & 6.67    & 7.09         & 1.83     & 5.28    & 8.8          & 6.74     & 8.68    & 8.76         & 5.07     & 7.74    & 8.94         & 1.75     & 4.08    & 8.67         & 4.11     & 6.99    & 7.95         \\
\cellcolor{close_source}{GPT-Image-1.5}~\cite{openai2025gpt-image-1.5}          & 7.75     & 8.65    & 3.16         & 5.37     & 6.34    & 3.29         & 5.34     & 4.36    & 1.72         & 4.29     & 5.61    & 3.21         & 2.97     & 5.75    & 5.54         & 6.50      & 7.37    & 3.34         & 7.54     & 7.70     & 3.04         & 2.51     & 4.47    & 5.09         & 5.44     & 6.34    & 3.41         \\
\cellcolor{close_source}{Seedream4.5~\cite{bytedance2025seedream4.5}}            & 6.33     & 7.56    & 5.88         & 5.01     & 6.14    & 5.17         & 6.25     & 6.51    & 4.23         & 3.82     & 5.61    & 4.37         & 3.04     & 5.14    & 6.53         & 7.27     & 8.07    & 6.22         & 6.51     & 7.73    & 6.42         & 1.85     & 4.55    & 7.49         & 5.10      & 6.43    & 5.56         \\ \midrule
\cellcolor{open_source}{OmniGen2~\cite{wu2025omnigen2}}               & 1.71     & 2.74    & 3.04         & 1.13     & 2.93    & 2.68         & 1.75     & 3.48    & 2.08         & 1.07     & 3.38    & 2.58         & 1.10     & 3.70    & 4.49         & 2.66     & 4.19    & 2.83         & 1.70     & 3.89    & 3.32         & 1.00     & 4.22    & 5.59         & 1.45     & 3.44    & 3.15         \\
\cellcolor{open_source}{BAGEL~\cite{deng2025bagel}}                  & 2.74     & 3.63    & 8.37         & 1.42     & 2.98    & 4.90         & 3.37     & 5.53    & 3.00         & 1.10     & 3.33    & 4.51         & 1.09     & 3.56    & 6.01         & 5.49     & 6.31    & 7.72         & 3.27     & 4.46    & 6.80         & 1.02     & 4.75    & 7.20         & 2.27     & 4.08    & 5.78         \\
\cellcolor{open_source}{Emu3.5~\cite{cui2025emu3.5}}                 & 5.69     & 7.03    & 3.16         & 2.99     & 4.04    & 2.50         & 4.88     & 4.18    & 1.91         & 1.36     & 3.36    & 2.14         & 1.47     & 2.96    & 4.95         & 5.51     & 6.00    & 3.96         & 5.76     & 6.36    & 3.31         & 1.12     & 1.82    & 3.81         & 3.68     & 4.60    & 3.06         \\
\cellcolor{open_source}{UniWorld-v2~\cite{li2025uniworldv2}}            & 5.03     & 6.98    & 5.94         & 2.51     & 4.00    & 3.86         & 4.11     & 5.59    & 2.25         & 1.30     & 3.55    & 2.90         & 1.36     & 2.86    & 6.15         & 5.63     & 6.41    & 5.26         & 4.62     & 6.03    & 4.96         & 1.13     & 3.35    & 7.62         & 3.18     & 4.84    & 4.55         \\
\cellcolor{open_source}{FLUX.2-dev~\cite{flux2}}             & 5.86     & 7.66    & 8.38         & 3.53     & 5.79    & 5.27         & 3.43     & 8.01    & 4.68         & 1.42     & \cellcolor{top}{6.03} & 5.58         & 1.40     & \cellcolor{top}{3.97} & 7.88         & 5.90     & 7.68    & 6.26         & 4.82     & 6.46    & 4.75         & \cellcolor{top}{1.34} & \cellcolor{top}{\textbf{5.63}} & \cellcolor{top}{\textbf{9.12}} & 3.42     & 6.36    & 6.26         \\
\cellcolor{open_source}{Qwen-Image-Edit-2511~\cite{qwen2025Qwen-Image-Edit-2511}}   & 5.51     & 7.41    & 8.05         & 3.11     & 4.75    & 6.17         & 5.90     & 7.44    & 5.65         & 1.54     & 4.32    & 5.70         & 1.56     & 2.92    & 8.28         & 6.51     & 7.44    & 8.02         & 5.24     & 6.92    & 7.50         & 1.19     & 4.64    & \cellcolor{top}{8.83} & 3.81     & 5.67    & 7.04         \\
\cellcolor{open_source}{LongCat-Image-Edit~\cite{team2025longcat}}     & 5.48     & 6.40    & 7.21         & 2.83     & 4.46    & 5.13         & 5.06     & 6.68    & 5.22         & 1.35     & 4.78    & 5.32         & 1.36     & 3.68    & 7.54         & 6.95     & 7.39    & 7.87         & 5.76     & 6.59    & 6.95         & 1.09     & \cellcolor{top}{5.34} & 7.80         & 3.68     & 5.52    & 6.39         \\
\cellcolor{open_source}{Step1X-Edit-v1.2~\cite{liu2025step1x}}       & 2.29     & 5.31    & 8.56         & 1.31     & 3.92    & 5.53         & 4.97     & 6.97    & 5.05         & 1.24     & 3.62    & 4.61         & 1.09     & 3.04    & 6.88         & 4.41     & 6.02    & 6.26         & 2.35     & 5.22    & 6.55         & 1.12     & \cellcolor{top}{5.26} & 7.97         & 2.26     & 4.77    & 6.29         \\
\cellcolor{open_source}{HY-image-3-instruct~\cite{cao2025hunyuanimage}}    & \cellcolor{top}{8.19} & \cellcolor{top}{\textbf{8.80}} & \cellcolor{top}{8.72} & \cellcolor{top}{4.33} & \cellcolor{top}{5.83} & \cellcolor{top}{6.50} & \cellcolor{top}{6.88} & 7.92    & \cellcolor{top}{5.82} & \cellcolor{top}{1.64} & 5.35    & \cellcolor{top}{7.06} & \cellcolor{top}{1.75} & 3.88    & \cellcolor{top}{8.66} & \cellcolor{top}{\textbf{8.05}} & \cellcolor{top}{\textbf{8.36}} & 8.20         & \cellcolor{top}{7.78} & \cellcolor{top}{\textbf{8.36}} & \cellcolor{top}{8.13} & 1.17     & 4.82    & 8.72         & \cellcolor{top}{5.03} & \cellcolor{top}{6.67} & \cellcolor{top}{7.58} \\
\cellcolor{open_source}{FireRed-Image-Edit~\cite{team2026firered}}     & 5.52     & 7.42    & 8.18         & 3.25     & 5.02    & \cellcolor{top}{6.53} & 6.57     & \cellcolor{top}{8.27} & 5.52         & \cellcolor{top}{1.71} & 4.91    & 5.98         & 1.62     & 3.78    & 8.47         & \cellcolor{top}{7.69} & 7.78    & \cellcolor{top}{8.31} & 5.86     & 7.26    & \cellcolor{top}{8.13} & 1.20     & 5.09    & \cellcolor{top}{9.00} & 4.12     & 6.14    & 7.29         \\ \midrule
\cellcolor{baseline}{Qwen-Image-Edit-2509~\cite{qwen2025Qwen-Image-Edit-2509}}   & 5.35     & 7.09    & 7.03         & 2.88     & 4.83    & 6.08         & 5.46     & 7.12    & 4.85         & 1.39     & 4.77    & 5.39         & 1.44     & 3.70    & 8.26         & 6.96     & 7.73    & 7.45         & 5.01     & 7.02    & 7.49         & 1.16     & 5.13    & 8.60         & 3.63     & 5.82    & 6.67         \\
\textbf{WeEdit-SFT (ours)}               & \cellcolor{top}{8.29} & \cellcolor{top}{8.50} & \cellcolor{top}{\textbf{9.25}} & \cellcolor{top}{5.52} & \cellcolor{top}{6.08} & \cellcolor{top}{8.33} & \cellcolor{top}{\textbf{8.60}} & \cellcolor{top}{\textbf{8.87}} & \cellcolor{top}{\textbf{8.17}} & \cellcolor{top}{\textbf{5.72}} & \cellcolor{top}{\textbf{6.63}} & \cellcolor{top}{8.07} & \cellcolor{top}{3.92} & \cellcolor{top}{4.08} & \cellcolor{top}{\textbf{8.77}} & \cellcolor{top}{7.98} & \cellcolor{top}{8.05} & \cellcolor{top}{\textbf{8.90}} & \cellcolor{top}{7.72} & \cellcolor{top}{7.71} & \cellcolor{top}{8.82} & \cellcolor{top}{1.80} & 4.88    & 8.55         & \cellcolor{top}{6.46} & \cellcolor{top}{6.98} & \cellcolor{top}{\textbf{8.58}} \\
 \textbf{WeEdit-RL (ours)}                & \cellcolor{top}{\textbf{8.47}} & \cellcolor{top}{8.33} & \cellcolor{top}{8.65} & \cellcolor{top}{\textbf{6.12}} & \cellcolor{top}{\textbf{6.30}} & \cellcolor{top}{\textbf{8.45}} & \cellcolor{top}{8.21} & \cellcolor{top}{8.76} & \cellcolor{top}{7.92} & \cellcolor{top}{5.58} & \cellcolor{top}{6.32} & \cellcolor{top}{\textbf{8.10}} & \cellcolor{top}{\textbf{5.13}} & \cellcolor{top}{\textbf{5.18}} & \cellcolor{top}{8.50} & 7.14     & \cellcolor{top}{8.15} & \cellcolor{top}{8.62} & \cellcolor{top}{\textbf{8.15}} & \cellcolor{top}{7.61} & \cellcolor{top}{\textbf{9.16}} & \cellcolor{top}{\textbf{2.54}} & 5.25    & 8.63         & \cellcolor{top}{\textbf{6.70}} & \cellcolor{top}{\textbf{7.10}} & \cellcolor{top}{8.49} \\
\textit{vs. \cellcolor{baseline}{Baseline}}            & \textbf{\green{+3.12}} & \textbf{\green{+1.24}} & \textbf{\green{+1.62}} & \textbf{\green{+3.24}} & \textbf{\green{+1.47}} & \textbf{\green{+2.37}} & \textbf{\green{+2.75}} & \textbf{\green{+1.64}} & \textbf{\green{+3.07}} & \textbf{\green{+4.19}} & \textbf{\green{+1.55}} & \textbf{\green{+2.71}} & \textbf{\green{+3.69}} & \textbf{\green{+1.48}} & \textbf{\green{+0.24}} & \textbf{\green{+0.18}} & \textbf{\green{+0.42}} & \textbf{\green{+1.17}} & \textbf{\green{+3.14}} & \textbf{\green{+0.59}} & \textbf{\green{+1.67}} & \textbf{\green{+1.38}} & \textbf{\green{+0.12}} & \textbf{\green{+0.03}} & \textbf{\green{+3.07}} & \textbf{\green{+1.28}} & \textbf{\green{+1.82}} \\ \bottomrule
\end{tabular}}
\label{tab: Quantitative Comparison Multilingual}
\vspace{-1.0em}
\end{table}

The Multilingual benchmark (\cref{tab: Quantitative Comparison Multilingual}) further reveals that WeEdit generalizes robustly to 15 diverse languages, including those with complex glyph systems such as Arabic, Thai, and Hindi. Our model achieves 6.70/7.10/8.49 overall, improving over the base model by +3.07/+1.28/+1.82. In contrast, several proprietary and open-source models suffer performance degradation when handling non-Latin scripts. This cross-lingual robustness can be attributed to our HTML-based data construction pipeline, which naturally scales to multilingual training data.

Finally, comparing WeEdit-SFT and WeEdit-RL reveals that the RL stage provides consistent and targeted improvements. On the Bilingual benchmark, the RL stage lifts overall IA from 6.99 to 7.47 and TC from 7.33 to 8.19. This confirms that the multi-dimensional reward design effectively steers the model toward higher editing quality beyond what pixel-level supervision alone can achieve.

\vspace{-1.0em}
\begin{figure}[t]
  \centering
  \includegraphics[width=1.0\linewidth]{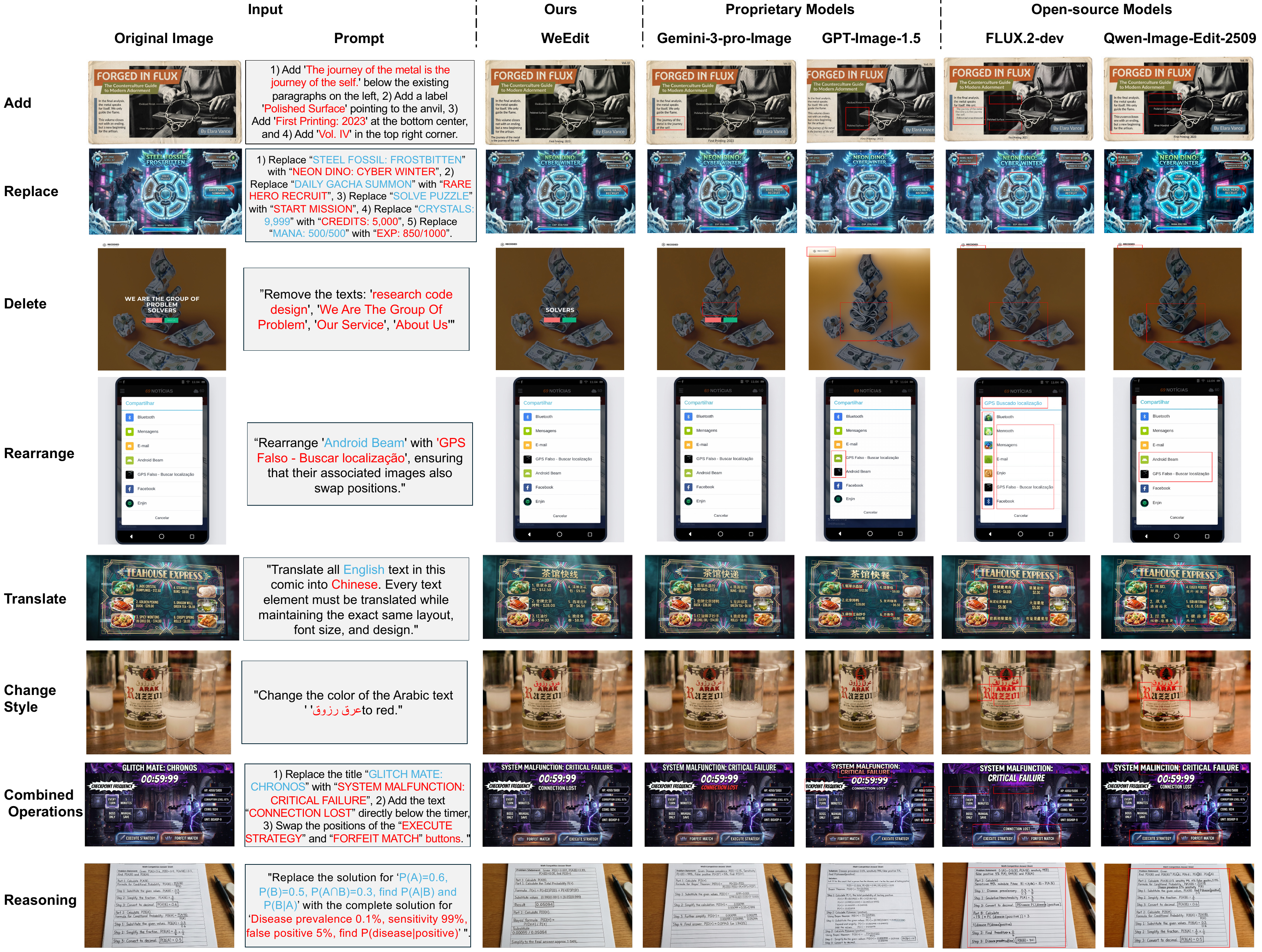}
  \caption{\textbf{Qualitative Results.} Regions with errors are highlighted in \textcolor{red}{red boxes}.}
  \label{fig:Qualitative_experiment}
  \vspace{-2.0em}
\end{figure}

\vspace{-0.5em}
\paragraph{\textbf{Qualitative Comparison.}}
\cref{fig:Qualitative_experiment} presents qualitative comparisons across all 8 editing operations. Open-source models such as FLUX.2-dev and Qwen-Image-Edit-2509 frequently produce blurry or hallucinated characters, especially when the instructions are complex or the target text is dense. The proprietary GPT-Image-1.5 encounters similar issues under challenging tasks. 
In contrast, WeEdit generates clear, accurately rendered text while faithfully executing instructions across all editing types, demonstrating robust performance on a wide range of challenging text-centric editing tasks. More detailed qualitative results are provided in the appendix.

\vspace{-0.5em}
\paragraph{\textbf{User Study.}}
To subjectively evaluate the editing quality, we conducted a user study comparing WeEdit against two open-source models and two proprietary systems. As illustrated in ~\cref{fig:user study}, human evaluators assessed the generated results across two dimensions: Instruction Adherence and Text Clarity on both Bilingual and Multilingual benchmarks. 
The results demonstrate that WeEdit significantly outperforms the open-source baselines Qwen-Image-Edit-2509 and FLUX.2-dev, as well as the proprietary GPT-Image-1.5, while achieving comparable performance with Gemini-3-Pro-Image.

\vspace{-0.75em}
\subsection{Ablation Study}
\vspace{-0.5em}
To systematically evaluate the contribution of each proposed module in WeEdit, we conduct a comprehensive ablation study, as summarized in \cref{tab:ablation}. 
As shown in the first row, the base model exhibits poor performance across 
three dimensions.
Applying glyph guidance directly to the baseline model at the inference stage (Row 2) results in negligible performance changes, indicating that the base model cannot effectively incorporate explicit spatial priors without specific training. 
Comparing the baseline with the SFT-only model (Row 3) reveals a significant performance leap across all three dimensions, which demonstrates the high quality and effectiveness of our constructed dataset. 
Furthermore, introducing RL directly on top of this SFT-only baseline (Row 4) struggles to break through to a higher performance ceiling. 
In contrast, combining SFT with Glyph guidance (Row 5) yields massive improvements, demonstrating a strong synergy where SFT aligns the model with the task while the Glyph prior provides the crucial spatial constraints needed for precise text rendering. 
Building upon this foundation, applying RL on top of the SFT-with-glyph model leads to further consistent enhancements. 
Within the RL stage, the design of the reward mechanism proves critical. 
Removing the Reference Image (RI) deprives the reward model of an explicit comparison anchor (Row 6), resulting in less accurate assessment of edit quality than when the reference is included.
Relying on a single reward model to score multiple dimensions simultaneously leads to sub-optimal performance due to metric entanglement (Row 7), whereas using Separate Reward Models (SRM) ensures the independent and accurate assessment of each dimension. 
The full WeEdit pipeline (Row 8) achieves the best results across all metrics, confirming the effectiveness of our training strategies and data.

\begin{figure}[t]
\centering
\begin{minipage}[b]{0.65\textwidth}
    \centering
    \includegraphics[width=\textwidth]{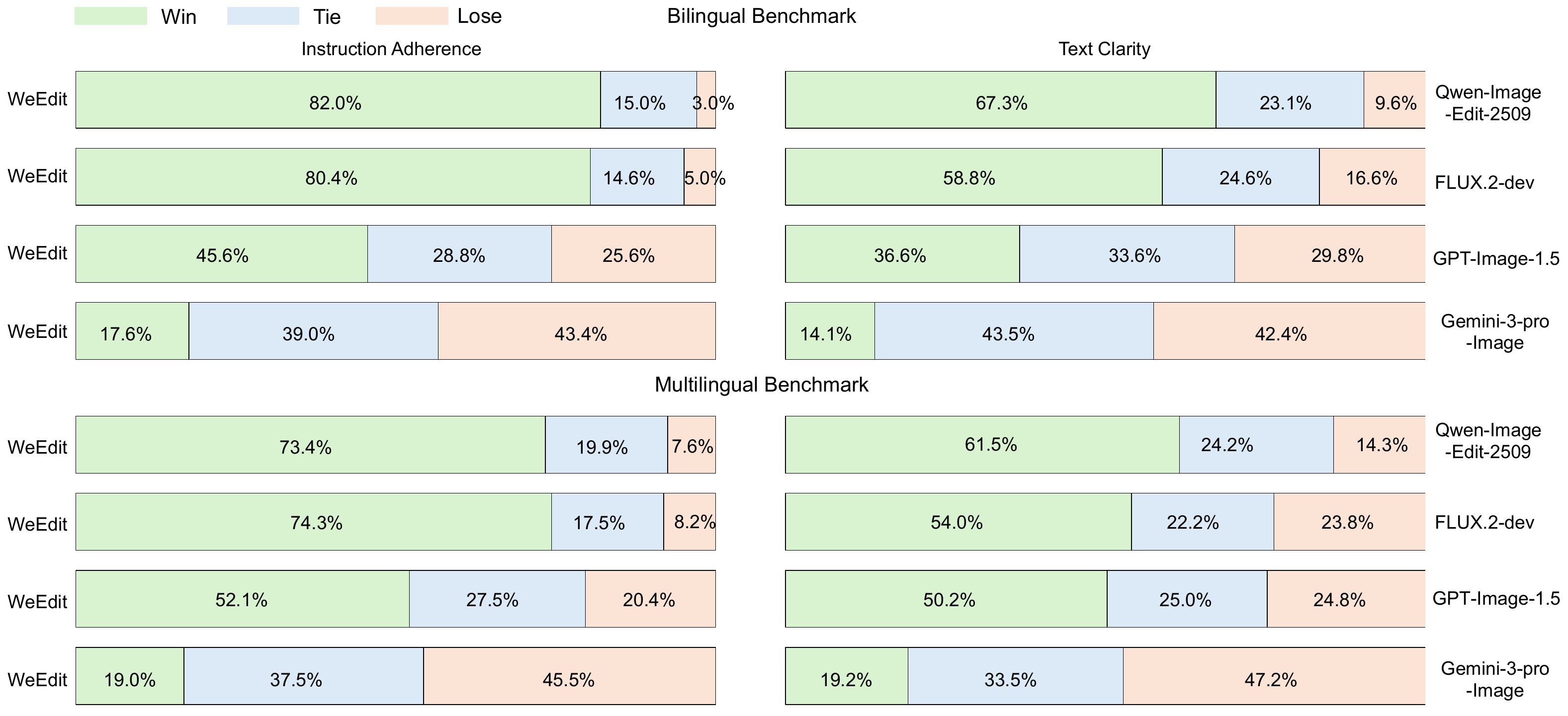}
    \caption{User Study.}
    \label{fig:user study}
    \par\vspace{0pt} 
\end{minipage}
\hfill
\begin{minipage}[b]{0.33\textwidth}
    \centering
    \tabcolsep=0.15cm
    \ra{1.1}
    
    \scalebox{0.54}{
    \begin{tabular}{@{}cccccccc@{}}
    \toprule
    \multicolumn{5}{c}{Ablation}            & \multicolumn{3}{c}{Performance} \\ \midrule
    SFT & Glyph & RL & RI & SRM & IA          & TC  & BP          \\ \midrule
        &       &    &           &          & 3.49        & 5.84        & 6.80         \\ 
        & $\checkmark$    &    &           &          & 3.58        & 4.67        & 6.82        \\ 
    $\checkmark$   &       &    &           &          &   5.32		           &  6.62           &  8.11           \\ 
    $\checkmark$   &       & $\checkmark$  & $\checkmark$         & $\checkmark$        &   5.41	   &   6.72          & 8.07            \\ 
    $\checkmark$   & $\checkmark$     &    &           &          & 6.99        & 7.33        & 8.63        \\ 
    $\checkmark$   & $\checkmark$     & $\checkmark$  &           & $\checkmark$        & 7.38		       & 7.91        & 8.92        \\ 
    $\checkmark$   & $\checkmark$     & $\checkmark$  & $\checkmark$         &          &    7.34         &    8.03         &  8.89           \\ 
    \rowcolor[HTML]{e9e9e9} 
    $\checkmark$   & $\checkmark$     & $\checkmark$  & $\checkmark$         & $\checkmark$        & \textbf{7.47}        & \textbf{8.19}        & \textbf{9.01}        \\ \bottomrule
    \end{tabular}}
    \caption{Ablation study on different modules in WeEdit.}
    \label{tab:ablation}
    \par\vspace{0pt} 
\end{minipage}
\vspace{-1.5em}
\end{figure}

\section{Conclusion}
\label{sec:conclusion}
In this paper, we presented WeEdit, a systematic solution for text-centric image editing that jointly addresses the challenges of model capability, data scarcity, and evaluation standardization. We introduced a glyph-guided supervised fine-tuning approach that leverages rendered glyph images as explicit spatial priors to enable precise text placement and character-level fidelity. We further proposed a multi-objective reinforcement learning stage with separate reward models and reference image grounding to optimize for instruction adherence, text readability, and background preservation. To overcome the lack of high-quality training data, we proposed a novel, scalable, HTML-based data construction pipeline that automatically synthesizes diverse editing pairs and naturally extends to multilingual settings. Additionally, we established a comprehensive benchmark covering diverse editing operations with bilingual and multilingual evaluation across 15 languages. Extensive experiments demonstrated that WeEdit surpasses all existing open-source models and most proprietary counterparts.

\clearpage 
\section*{Acknowledgments}
This work is supported by the National Natural Science Foundation of China (Grant No. 62472098) and the Science and Technology Commission of Shanghai Municipality (No. 25511106100).

\bibliographystyle{splncs04}
\bibliography{main}

\clearpage 

\section{Appendix}

 \begin{figure}[ht]
  \centering
  \includegraphics[width=1.0\linewidth]{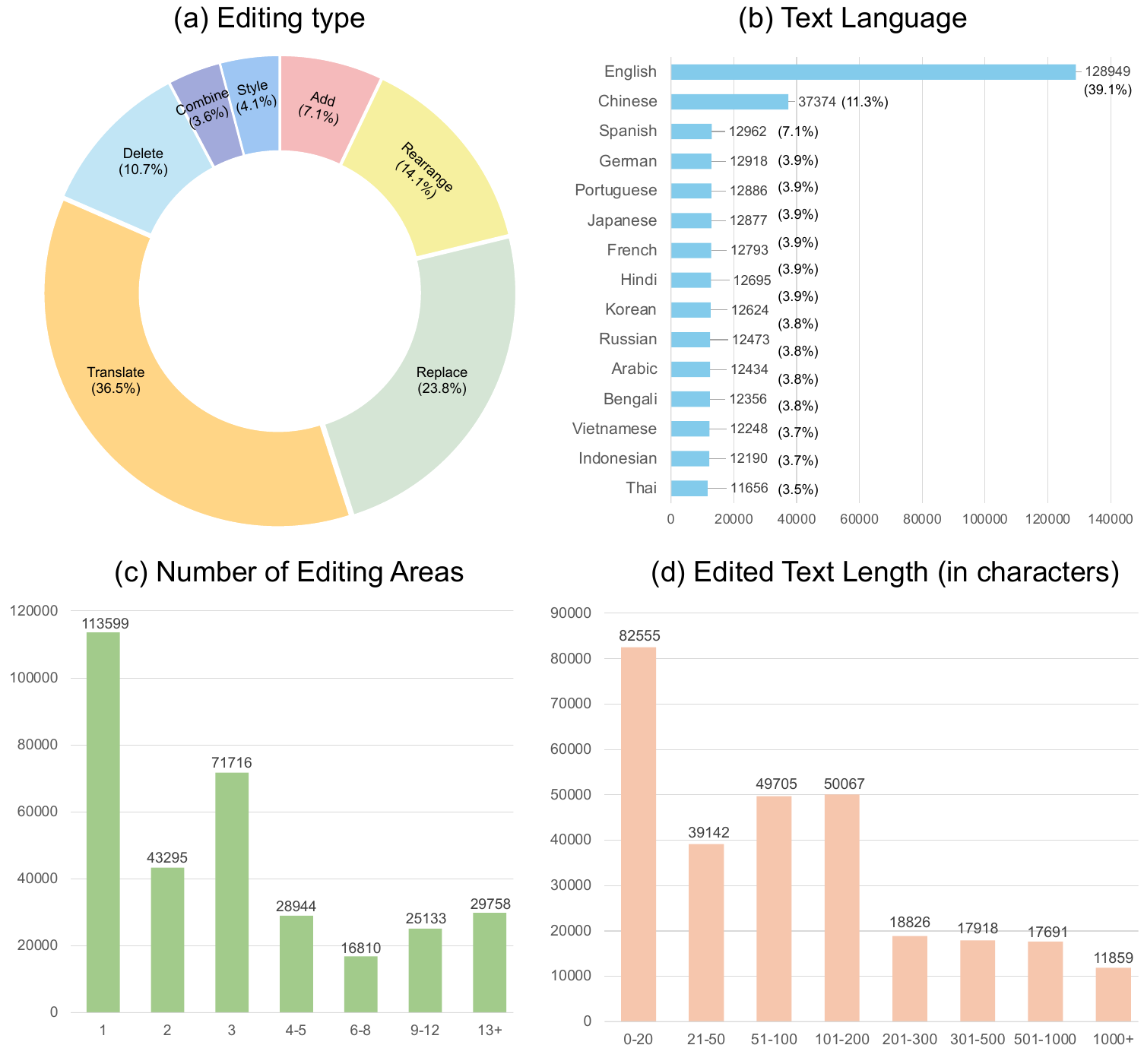}
  \caption{Statistics of WeEdit Dataset: (a) Distribution over the seven editing operation types. (b) Language distribution across 15 supported languages. (c) Distribution of the number of edited regions per sample. (d) Distribution of the total edited text length (in characters) per sample.
  }
  \label{fig:data_statistics}
\end{figure}

\subsection{Datasets}

\paragraph{\textbf{Statistics of Datasets}}

We construct a training set containing 330K samples, including approximately 160K unstructured text pairs and 170K structured text pairs. The dataset spans 7 editing operation types and covers 15 languages.
As shown in~\cref{fig:data_statistics}(a), translation (36.5\%) and replacement (23.8\%) together account for over 60\% of all samples, reflecting their prevalence in real-world text editing scenarios. Rearrange (14.1\%) and delete (10.7\%) operations also constitute substantial portions, while add (7.1\%), style change (4.1\%), and combined editing (3.6\%) provide complementary coverage of less frequent but practically important operations.
\cref{fig:data_statistics}(b) illustrates the language distribution. Although English (39.1\%) and Chinese (11.3\%) dominate due to their abundance in existing corpora, we intentionally balance the remaining 13 languages---including Spanish, German, Portuguese, Japanese, French, Hindi, Korean, Russian, Arabic, Bengali, Vietnamese, Indonesian, and Thai---to each comprise approximately 3.5\%--7.1\% of the dataset, ensuring broad multilingual coverage.
In terms of editing complexity, \cref{fig:data_statistics}(c) shows that single-region edits are the most common (113{,}599 samples), yet a significant proportion of samples involve multi-region editing: over 100K samples require editing two or more regions simultaneously, and nearly 30K samples involve 13 or more regions. This distribution encourages the model to learn both precise localized edits and coordinated multi-region modifications.
Finally, \cref{fig:data_statistics}(d) reveals a wide spread in edited text length. While short edits (0--20 characters) are the most frequent (82{,}555 samples), a long tail extends to over 1{,}000 characters (11{,}859 samples), covering scenarios from brief label modifications to full paragraph-level rewrites. 
This diversity in both region count and text length ensures that our dataset comprehensively captures the complexity spectrum of real-world text image editing tasks.

\subsection{Experimental Results}

\paragraph{\textbf{Cross-Judge Reliability of the Bilingual Benchmark}}

Since our evaluation protocol relies on a single VLM judge~(Gemini-3-Pro) to score edited images, we verify the robustness of our conclusions by re-evaluating the entire Bilingual Benchmark using a cross-family judge, GPT-5.4, and comparing its scores against those of Gemini-3-Pro.

As reported in \cref{tab:cross_judge}, the two judges demonstrate a remarkably high degree of consistency. Across all $9\times8\times3=243$ per-cell scores, their ratings align closely. More importantly, the two judges produce an \textit{identical} Overall-IA ranking of all 9 models, with WeEdit remaining the strongest open-source model and trailing only the proprietary Gemini-3-Image. This strong agreement across judges from different model families indicates that the benchmark rankings reflect genuine differences in editing quality rather than judge-specific bias.

\begin{table}[ht]
\centering
\caption{\textbf{Quantitative Results on the Bilingual Benchmark evaluated by the cross-family judge GPT-5.4.} We compare WeEdit with \colorbox{close_source}{proprietary} and \colorbox{open_source}{open-source} models across 8 editing operations. IA, TC, and BP denote Instruction Adherence, Text Clarity, and Background Preservation. The best open-source results are in \textbf{bold}, with \colorbox{top}{top-3} highlighted. The resulting rankings are consistent with those obtained under the Gemini-3-Pro judge.}
\label{tab:cross_judge}
\tabcolsep=0.03cm
\ra{1.1}
\scalebox{0.4}{
\begin{tabular}{@{}lccc|ccc|ccc|ccc|ccc|ccc|ccc|ccc|ccc@{}}
\toprule                           
\multirow{2}{*}{\textbf{Model}}            & \multicolumn{3}{c|}{\textbf{Add}}          & \multicolumn{3}{c|}{\textbf{Replace}}      & \multicolumn{3}{c|}{\textbf{Delete}}       & \multicolumn{3}{c|}{\textbf{Rearrange}}    & \multicolumn{3}{c|}{\textbf{Translate}}    & \multicolumn{3}{c|}{\textbf{Style}} & \multicolumn{3}{c|}{\textbf{Combined}}     & \multicolumn{3}{c|}{\textbf{Reasoning}}    & \multicolumn{3}{c}{\textbf{Overall}}      \\ \cmidrule(l){2-28}
                                  & IA & TC & BP & IA & TC & BP & IA & TC & BP & IA & TC & BP & IA & TC & BP & IA & TC & BP & IA & TC & BP & IA & TC & BP & IA & TC & BP \\ \midrule
\cellcolor{close_source}{Gemini-3-Image} & \textbf{8.16} & \textbf{7.97} & \textbf{8.61} & \textbf{8.08} & \textbf{7.75} & \textbf{8.30} & \textbf{7.35} & \textbf{7.76} & \textbf{7.58} & \textbf{7.88} & \textbf{7.99} & \textbf{8.41} & \textbf{6.73} & \textbf{7.92} & \textbf{8.79} & \textbf{8.56} & \textbf{8.46} & 8.52 & \textbf{8.66} & \textbf{8.11} & \textbf{8.48} & 4.18 & \textbf{7.54} & \textbf{7.81} & \textbf{7.66} & \textbf{7.98} & \textbf{8.34} \\
\cellcolor{close_source}{Gemini-2.5-Image} & 6.32 & 7.23 & 7.54 & 4.35 & 6.24 & 7.79 & 6.02 & 7.28 & 6.53 & 2.55 & 7.18 & 7.35 & 2.29 & 6.31 & 8.19 & 7.67 & 8.36 & \textbf{8.95} & 5.45 & 6.83 & 7.98 & 2.40 & 5.73 & 7.70 & 4.58 & 6.88 & 7.64 \\
\cellcolor{close_source}{GPT-Image-1.5} & 7.30 & 7.83 & 6.59 & 6.67 & 7.26 & 5.92 & 6.03 & 6.26 & 4.30 & 5.94 & 7.12 & 5.93 & 4.72 & 7.40 & 7.83 & 8.11 & 8.31 & 7.35 & 7.63 & 7.77 & 6.19 & \textbf{4.19} & 6.02 & 7.11 & 6.42 & 7.33 & 6.23 \\ \midrule
\cellcolor{open_source}{FLUX.2-dev} & 5.65 & 6.96 & 5.60 & 4.42 & 6.17 & 5.88 & 3.89 & 6.72 & 5.68 & 2.30 & \cellcolor{top}{6.18} & \cellcolor{top}{6.17} & 1.80 & \cellcolor{top}{5.06} & \cellcolor{top}{8.21} & 7.23 & 7.97 & 7.67 & 5.31 & 6.59 & 5.60 & 1.61 & 4.51 & 6.41 & 3.97 & \cellcolor{top}{6.38} & 6.36 \\
\cellcolor{open_source}{Step1X-Edit-v1.2} & 3.68 & 4.43 & 6.83 & 3.30 & 4.11 & \cellcolor{top}{6.78} & 5.43 & 5.91 & 6.09 & 2.19 & 4.21 & 4.97 & 1.87 & 4.14 & 7.49 & 7.19 & 7.77 & 8.08 & 3.30 & 4.68 & 6.92 & 1.48 & 4.37 & 6.29 & 3.43 & 4.85 & 6.68 \\
\cellcolor{open_source}{FireRed-Image-Edit} & \cellcolor{top}{6.18} & \cellcolor{top}{7.10} & \cellcolor{top}{7.54} & \cellcolor{top}{5.15} & \cellcolor{top}{6.23} & 6.75 & \cellcolor{top}{6.87} & \cellcolor{top}{7.42} & \cellcolor{top}{6.38} & \cellcolor{top}{2.70} & 5.50 & 5.93 & \cellcolor{top}{2.25} & 4.68 & 7.94 & \cellcolor{top}{8.27} & \cellcolor{top}{8.20} & \cellcolor{top}{8.67} & \cellcolor{top}{5.60} & \cellcolor{top}{6.87} & \cellcolor{top}{7.05} & 1.63 & \cellcolor{top}{4.67} & \cellcolor{top}{6.88} & \cellcolor{top}{4.82} & 6.37 & \cellcolor{top}{7.09} \\ \midrule
\cellcolor{baseline}{Qwen-Edit-2509} & 5.67 & 6.73 & 6.99 & 4.23 & 5.57 & 6.37 & 5.45 & 6.66 & 5.51 & 2.25 & 5.59 & 6.07 & 2.02 & 4.17 & 8.03 & 7.96 & 8.14 & 8.26 & 5.13 & 6.25 & 6.82 & \cellcolor{top}{1.64} & \cellcolor{top}{4.59} & 6.81 & 4.18 & 5.93 & 6.78 \\
\textbf{WeEdit-SFT (ours)} & \cellcolor{top}{7.36} & \cellcolor{top}{7.32} & \cellcolor{top}{8.42} & \cellcolor{top}{6.72} & \cellcolor{top}{6.71} & \cellcolor{top}{8.26} & \cellcolor{top}{\textbf{8.16}} & \cellcolor{top}{7.92} & \cellcolor{top}{\textbf{7.93}} & \cellcolor{top}{7.11} & \cellcolor{top}{7.07} & \cellcolor{top}{8.31} & \cellcolor{top}{5.87} & \cellcolor{top}{5.98} & \cellcolor{top}{8.48} & \cellcolor{top}{\textbf{8.40}} & \cellcolor{top}{8.27} & \cellcolor{top}{8.64} & \cellcolor{top}{7.49} & \cellcolor{top}{7.30} & \cellcolor{top}{8.32} & \cellcolor{top}{3.32} & 4.52 & \cellcolor{top}{7.01} & \cellcolor{top}{7.00} & \cellcolor{top}{6.94} & \cellcolor{top}{8.24} \\
\textbf{WeEdit-RL (ours)} & \cellcolor{top}{\textbf{7.68}} & \cellcolor{top}{\textbf{7.63}} & \cellcolor{top}{\textbf{8.63}} & \cellcolor{top}{\textbf{7.13}} & \cellcolor{top}{\textbf{7.29}} & \cellcolor{top}{\textbf{8.66}} & \cellcolor{top}{8.08} & \cellcolor{top}{\textbf{8.01}} & \cellcolor{top}{7.69} & \cellcolor{top}{\textbf{7.83}} & \cellcolor{top}{\textbf{7.81}} & \cellcolor{top}{\textbf{8.69}} & \cellcolor{top}{\textbf{6.57}} & \cellcolor{top}{\textbf{7.09}} & \cellcolor{top}{\textbf{8.84}} & \cellcolor{top}{8.34} & \cellcolor{top}{\textbf{8.31}} & \cellcolor{top}{\textbf{8.76}} & \cellcolor{top}{\textbf{7.89}} & \cellcolor{top}{\textbf{7.75}} & \cellcolor{top}{\textbf{8.41}} & \cellcolor{top}{\textbf{4.10}} & \cellcolor{top}{\textbf{5.27}} & \cellcolor{top}{\textbf{7.74}} & \cellcolor{top}{\textbf{7.39}} & \cellcolor{top}{\textbf{7.51}} & \cellcolor{top}{\textbf{8.46}} \\
\textit{vs. \cellcolor{baseline}{Baseline}} & \small{\textbf{\green{+2.01}}} & \small{\textbf{\green{+0.90}}} & \small{\textbf{\green{+1.64}}} & \small{\textbf{\green{+2.90}}} & \small{\textbf{\green{+1.72}}} & \small{\textbf{\green{+2.29}}} & \small{\textbf{\green{+2.63}}} & \small{\textbf{\green{+1.35}}} & \small{\textbf{\green{+2.18}}} & \small{\textbf{\green{+5.58}}} & \small{\textbf{\green{+2.22}}} & \small{\textbf{\green{+2.62}}} & \small{\textbf{\green{+4.55}}} & \small{\textbf{\green{+2.92}}} & \small{\textbf{\green{+0.81}}} & \small{\textbf{\green{+0.38}}} & \small{\textbf{\green{+0.17}}} & \small{\textbf{\green{+0.50}}} & \small{\textbf{\green{+2.76}}} & \small{\textbf{\green{+1.50}}} & \small{\textbf{\green{+1.59}}} & \small{\textbf{\green{+2.46}}} & \small{\textbf{\green{+0.68}}} & \small{\textbf{\green{+0.93}}} & \small{\textbf{\green{+3.21}}} & \small{\textbf{\green{+1.58}}} & \small{\textbf{\green{+1.68}}} \\ \bottomrule
\end{tabular}}
\end{table}

\paragraph{\textbf{More Qualitative Results}}

We conduct a detailed qualitative comparisons of WeEdit against six representative methods---Gemini-3-pro-Image, GPT-Image-1.5, FLUX.2-dev, Qwen-Image-Edit-2509, FireRed-Image-Edit, and Step1X-Edit-v1.2---across eight editing operation types in \cref{fig:qualitative_add,fig:qualitative_replace,fig:qualitative_delete_rearrange,fig:qualitative_translate,fig:qualitative_style,fig:qualitative_combined,fig:qualitative_reasoning}. We annotate three common failure modes: \textcolor{my_red}{red} boxes for inaccurate instruction execution, \textcolor{my_purple}{purple} boxes for unclear or illegible rendered text, and \textcolor{my_orange}{orange} boxes for unintended modifications to non-edited regions. Several key observations emerge from these comparisons.
\textbf{First}, existing methods frequently fail to execute all sub-instructions faithfully, particularly when multiple editing operations are specified simultaneously. For instance, in the \textit{add} task (\cref{fig:qualitative_add}), most baselines omit at least one of the four required additions, whereas WeEdit accurately places all specified text elements at the correct locations. \textbf{Second}, text clarity remains a significant challenge for open-source models. As shown in the \textit{replace} (\cref{fig:qualitative_replace}) and \textit{translate} (\cref{fig:qualitative_translate}) examples, methods such as FireRed-Image-Edit and Step1X-Edit-v1.2 frequently produce blurry or garbled characters, especially for non-Latin scripts~(\eg Chinese). In contrast, WeEdit consistently renders sharp and legible text across all languages. \textbf{Third}, background preservation is a persistent weakness of existing approaches. Both proprietary models (\eg, GPT-Image-1.5) and open-source methods (\eg, FLUX.2-dev) often introduce noticeable artifacts or alter the visual appearance of non-edited regions, as evident in the \textit{delete} and \textit{style change} tasks (\cref{fig:qualitative_delete_rearrange,fig:qualitative_style}). WeEdit, benefiting from its task-aware reward design that explicitly penalizes background degradation, maintains high fidelity in unedited areas. \textbf{Fourth}, for complex operations such as \textit{rearrange} (\cref{fig:qualitative_delete_rearrange}), \textit{combined editing} (\cref{fig:qualitative_combined}), and \textit{reasoning-based editing} (\cref{fig:qualitative_reasoning}), most baselines either misinterpret the spatial semantics of the instruction or fail to produce logically coherent content.
These results collectively demonstrate that WeEdit achieves superior performance in instruction adherence, text clarity, background preservation, and semantic reasoning across diverse editing scenarios.

\subsection{Limitations}
While WeEdit demonstrates SOTA performance across a wide range of visual text editing tasks, it still possesses a few limitations that provide avenues for future research. First, similar to other diffusion-based generation models, rendering extremely dense or tiny text in non-ultra-high-resolution images can sometimes result in minor blurring or structural artifacts. Second, while WeEdit exhibits promising capabilities in reasoning-aware text editing, the underlying multimodal large language model may still occasionally produce factual hallucinations or logical errors when confronted with extremely complex, domain-specific reasoning tasks. Future work will focus on further exploring the boundary capabilities for dense and ultra-small text rendering, as well as integrating more robust reasoning modules to mitigate logical errors in complex scenarios.

\begin{figure}[ht]
  \centering
  \includegraphics[width=1.0\linewidth]{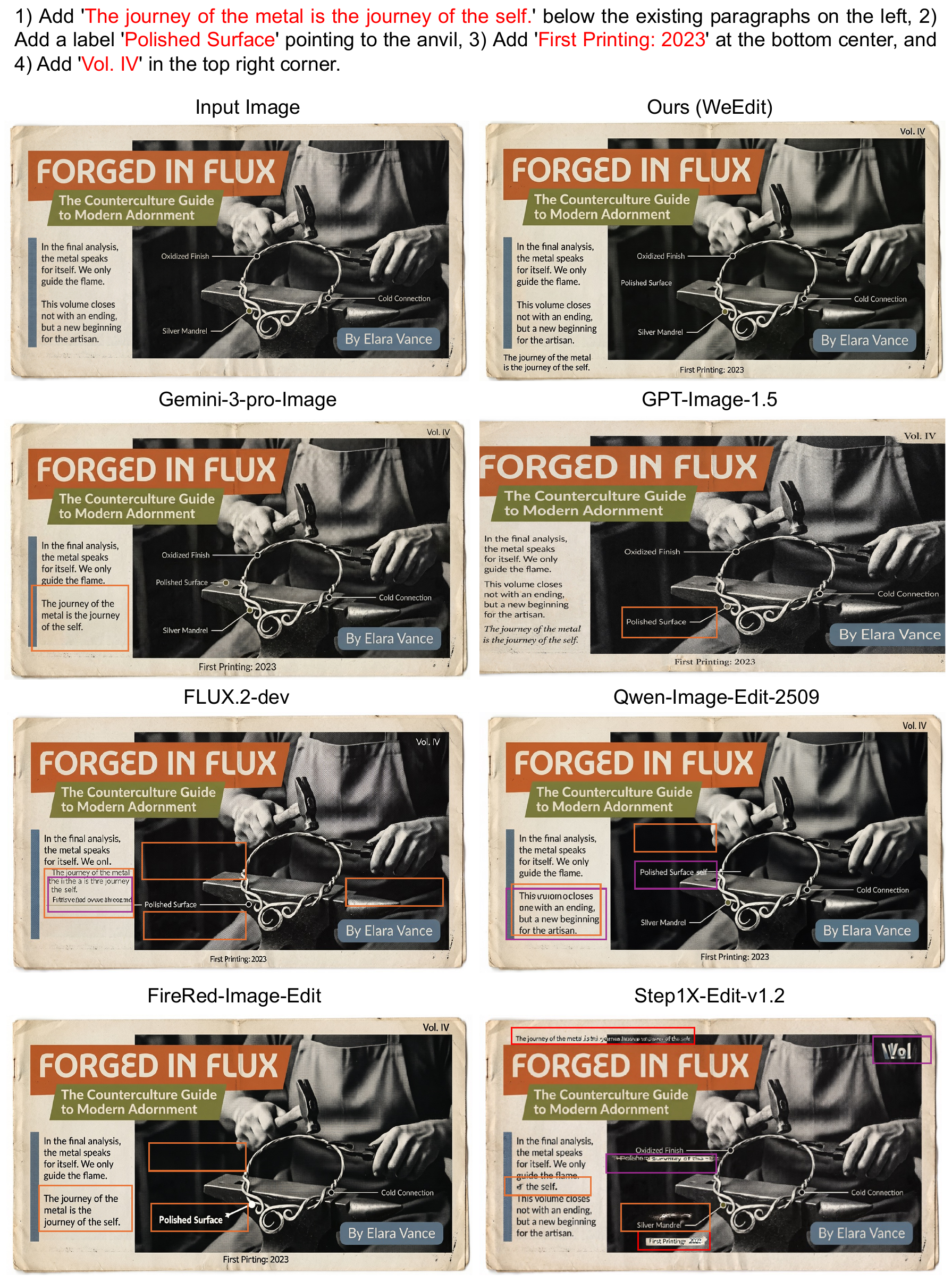}
  \caption{Qualitative comparison of the \textbf{add} operation. Inaccurate instruction execution is highlighted with \colorbox{my_box_red}{red}boxes, unclear text rendering with \colorbox{my_box_purple}{purple}boxes, and unintended alterations to non-editing regions with \colorbox{my_box_orange}{orange}boxes.
  }
  \label{fig:qualitative_add}
\end{figure}

 \begin{figure}[ht]
  \centering
  \includegraphics[width=1.0\linewidth]{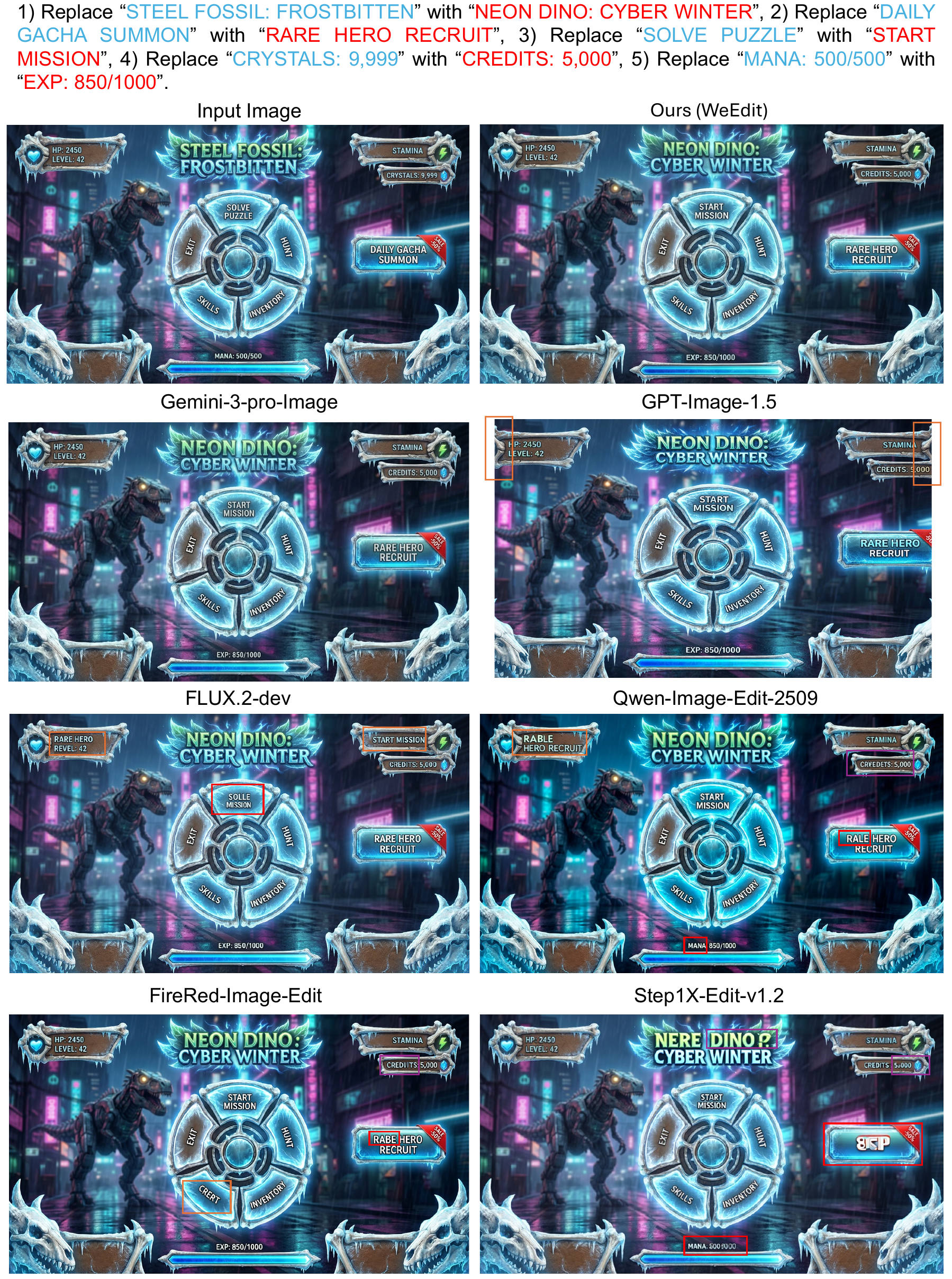}
  \caption{Qualitative comparison of the \textbf{replace} operation. Inaccurate instruction execution is highlighted with \colorbox{my_box_red}{red}boxes, unclear text rendering with \colorbox{my_box_purple}{purple}boxes, and unintended alterations to non-editing regions with \colorbox{my_box_orange}{orange}boxes.
  }
  \label{fig:qualitative_replace}
\end{figure}

 \begin{figure}[ht]
  \centering
  \includegraphics[width=1.0\linewidth]{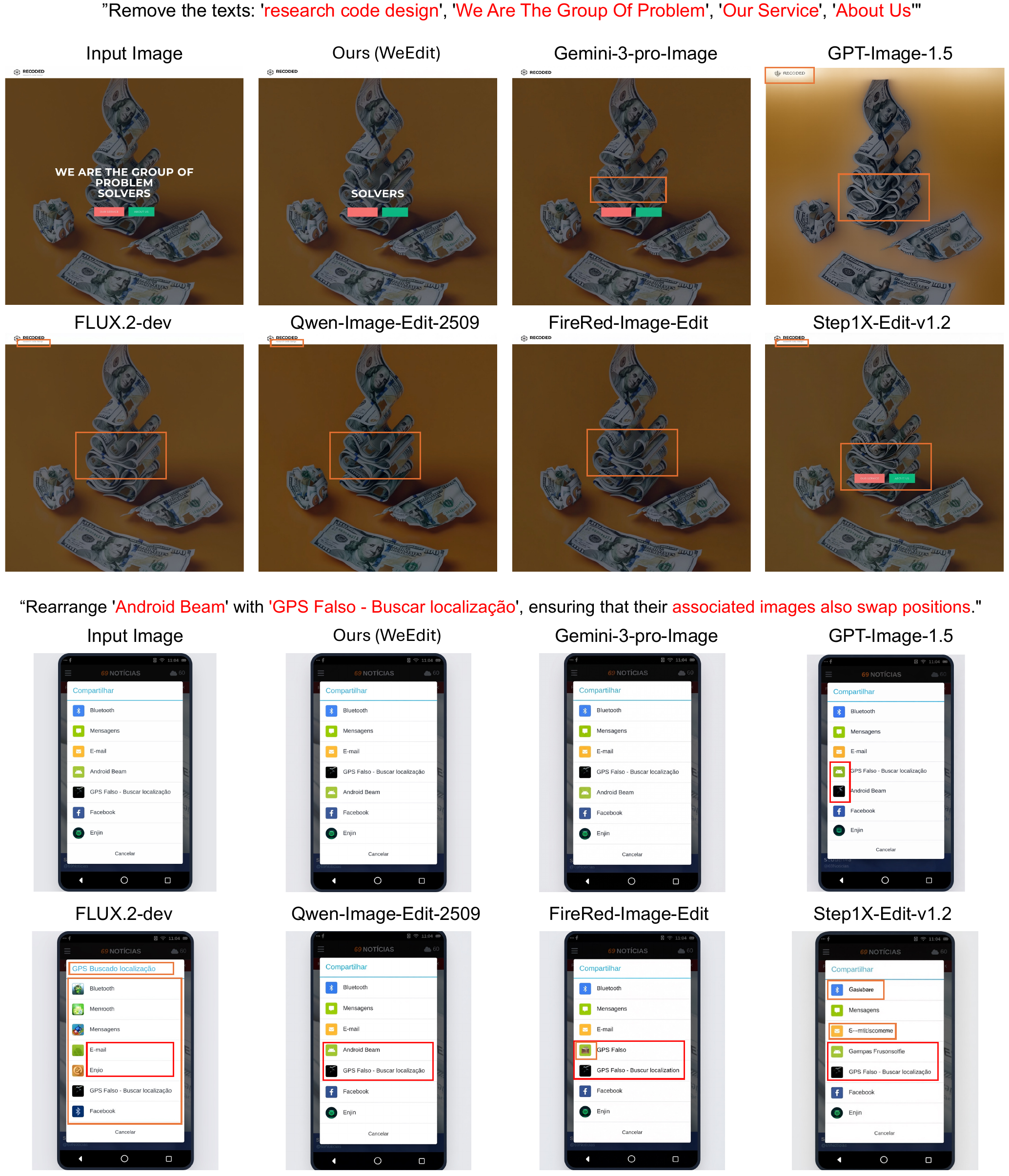}
  \caption{Qualitative comparison of the \textbf{delete} and \textbf{rearrange} operations. Inaccurate instruction execution is highlighted with \colorbox{my_box_red}{red}boxes, unclear text rendering with \colorbox{my_box_purple}{purple}boxes, and unintended alterations to non-editing regions with \colorbox{my_box_orange}{orange}boxes.
  }
  \label{fig:qualitative_delete_rearrange}
\end{figure}

 \begin{figure}[ht]
  \centering
  \includegraphics[width=1.0\linewidth]{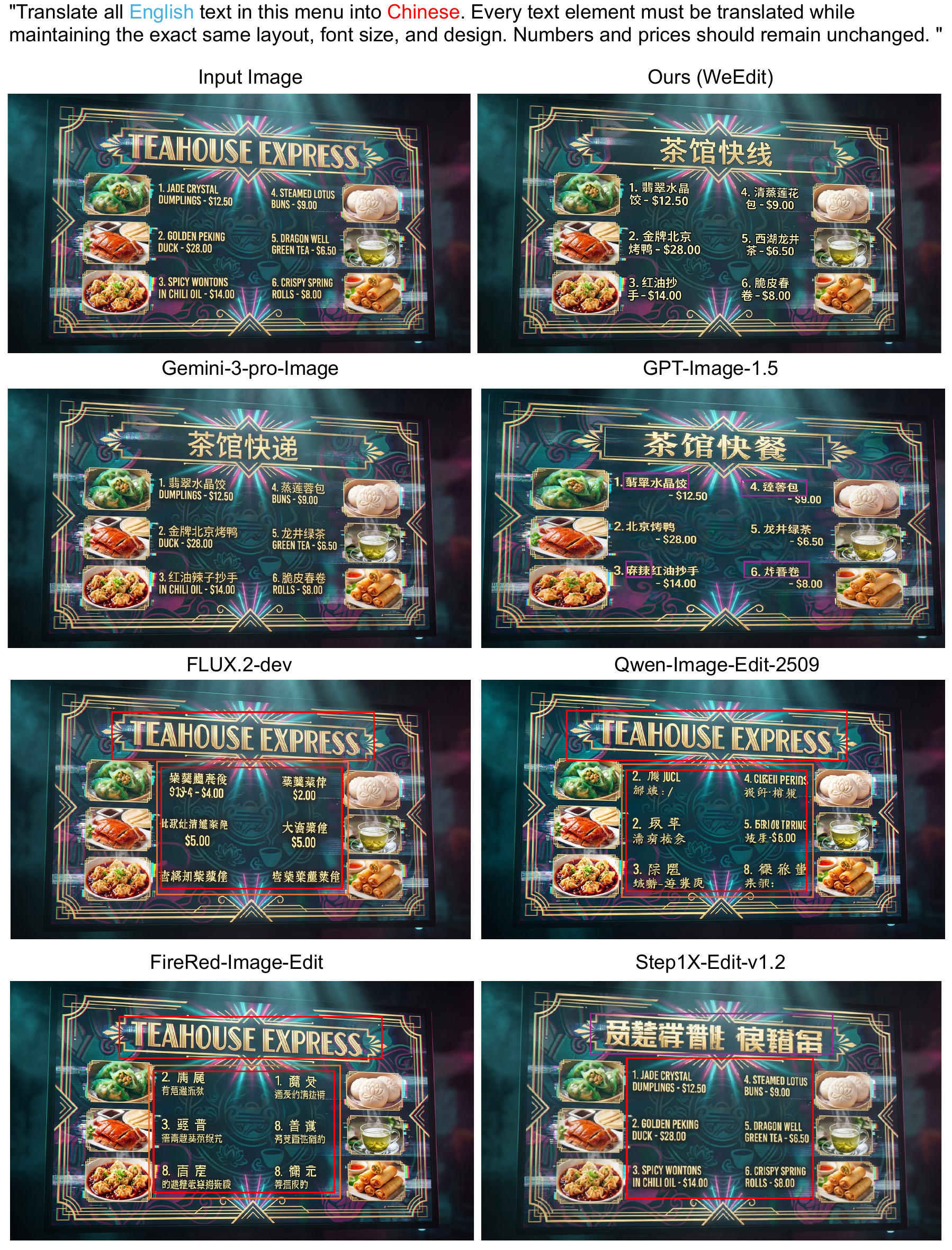}
  \caption{Qualitative comparison of the \textbf{translate} operation. Inaccurate instruction execution is highlighted with \colorbox{my_box_red}{red}boxes, unclear text rendering with \colorbox{my_box_purple}{purple}boxes, and unintended alterations to non-editing regions with \colorbox{my_box_orange}{orange}boxes.
  }
  \label{fig:qualitative_translate}
\end{figure}

 \begin{figure}[ht]
  \centering
  \includegraphics[width=1.0\linewidth]{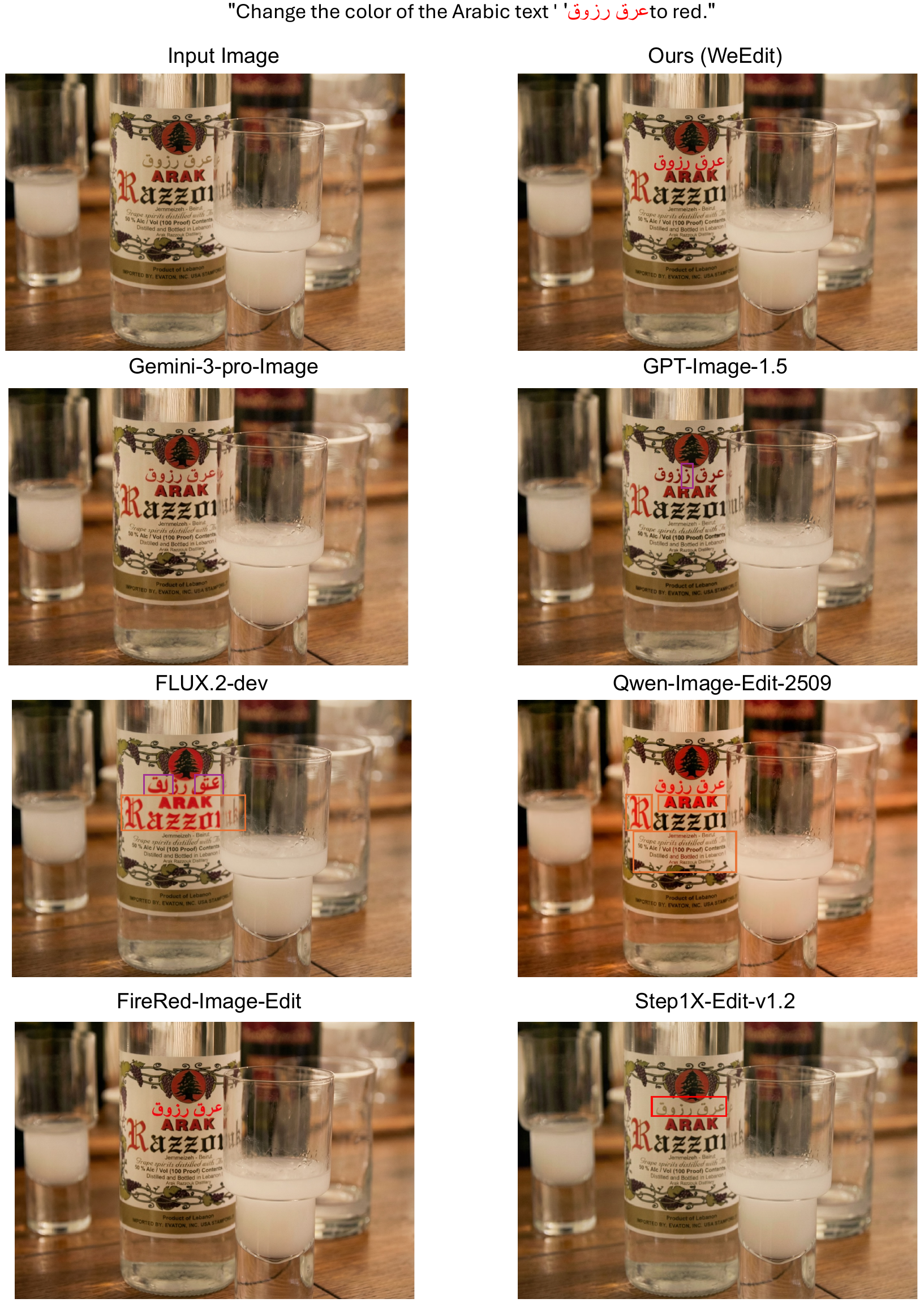}
  \caption{Qualitative comparison of the \textbf{change style} operation. Inaccurate instruction execution is highlighted with \colorbox{my_box_red}{red}boxes, unclear text rendering with \colorbox{my_box_purple}{purple}boxes, and unintended alterations to non-editing regions with \colorbox{my_box_orange}{orange}boxes.
  }
  \label{fig:qualitative_style}
\end{figure}

 \begin{figure}[ht]
  \centering
  \includegraphics[width=1.0\linewidth]{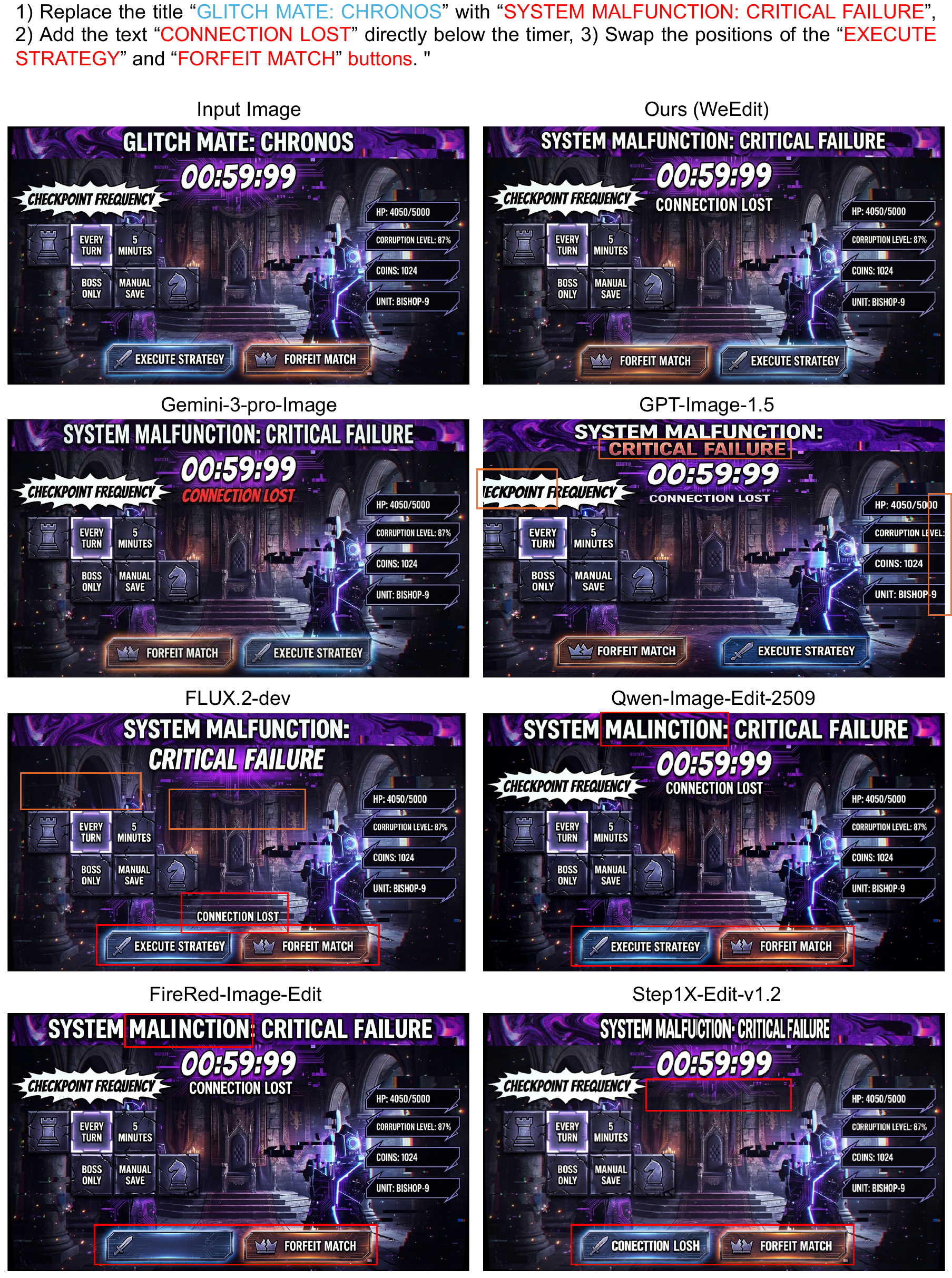}
  \caption{Qualitative comparison of \textbf{combined} operations. Inaccurate instruction execution is highlighted with \colorbox{my_box_red}{red}boxes, unclear text rendering with \colorbox{my_box_purple}{purple}boxes, and unintended alterations to non-editing regions with \colorbox{my_box_orange}{orange}boxes.
  }
  \label{fig:qualitative_combined}
\end{figure}

 \begin{figure}[ht]
  \centering
  \includegraphics[width=1.0\linewidth]{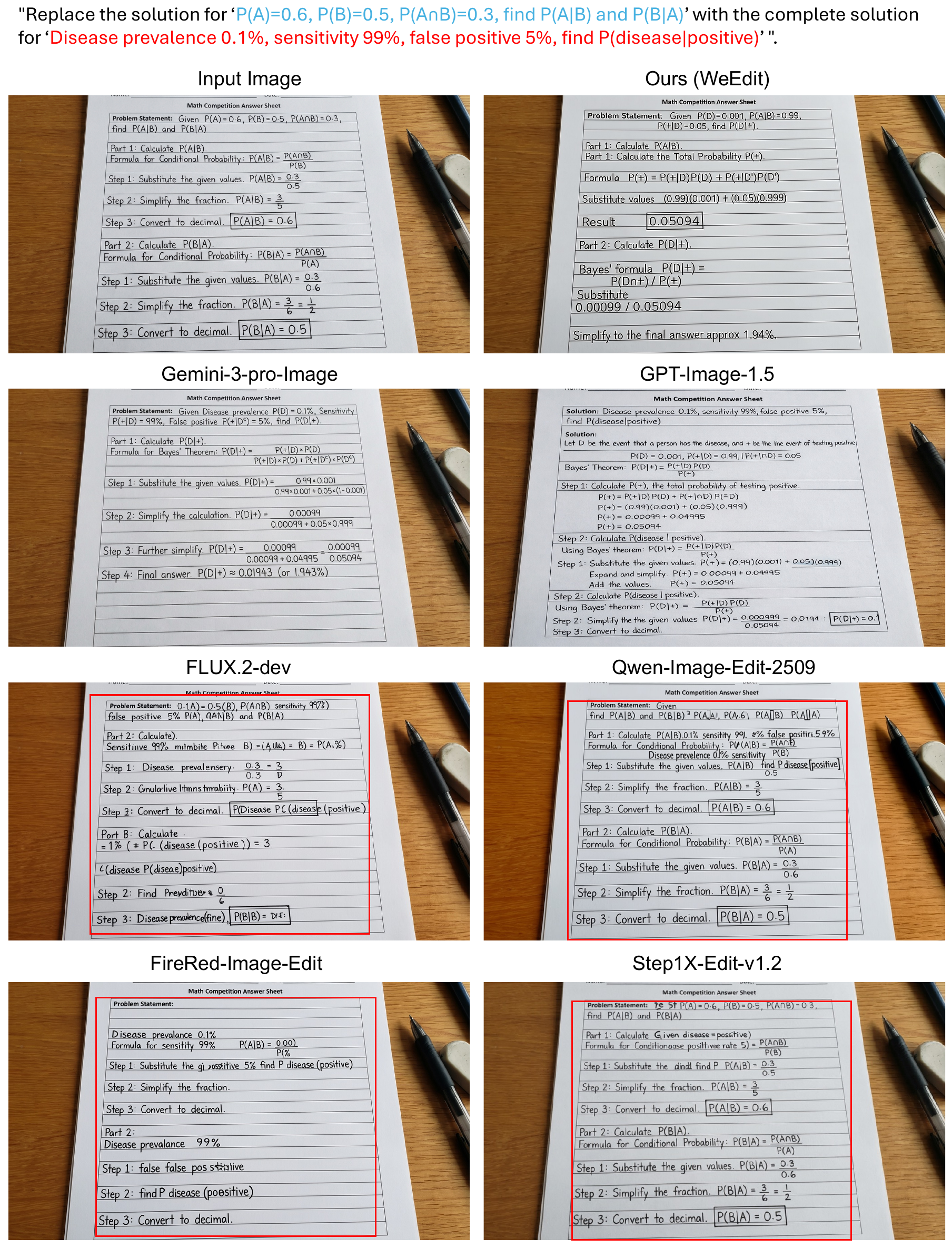}
  \caption{Qualitative comparison of the \textbf{reasoning} operation. 
  Inaccurate instruction execution is highlighted with \colorbox{my_box_red}{red}boxes.
  }
  \label{fig:qualitative_reasoning}
\end{figure}

\end{document}